\documentclass{article}
\usepackage[numbers]{natbib}
\usepackage[preprint]{neurips_2020}
\usepackage[utf8]{inputenc} 
\usepackage[T1]{fontenc}    
\usepackage{titlesec}
\usepackage{hyperref}       
\usepackage{url}            
\usepackage{booktabs}       
\usepackage{amsfonts}       
\usepackage{nicefrac}       
\usepackage{microtype}      
\usepackage{xcolor}
\usepackage{subcaption}
\usepackage{graphicx}
\usepackage{enumitem}
\usepackage[export]{adjustbox}
\usepackage{amsmath}
\usepackage{multirow}

\usepackage{wrapfig,booktabs}

\title{NAS-Bench-NLP: Neural Architecture Search Benchmark for Natural Language Processing}

\author{%
  Nikita Klyuchnikov$^{1}$\thanks{corresponding author, e-mail: \texttt{nikita.klyuchnikov@skolkovotech.ru}} \\
  \And
  Ilya Trofimov$^{1}$ \\
  \And
  Ekaterina Artemova$^{2}$ \\
  \AND
  Mikhail Salnikov$^{1}$ \\
  \And
  Maxim Fedorov$^{1}$ \\
  \And
  Evgeny Burnaev$^{1}$ \\
  \AND
  \\ 
  $^{1}$\text{Skolkovo Institute of Science and Technology},\\ $^{2}$\text{National Research University Higher School of Economics}\\
}

\begin{document}

\maketitle
\begin{abstract}

Neural Architecture Search (NAS) is a promising and rapidly evolving research area. Training a large number of neural networks requires an exceptional amount of computational power, which makes NAS unreachable for those researchers who have limited or no access to high-performance clusters and supercomputers. A few benchmarks with precomputed neural architectures performances have been recently introduced to overcome this problem and ensure more reproducible experiments. However, these benchmarks are only for the computer vision domain and, thus, are built from the image datasets and convolution-derived architectures. In this work, we step outside the computer vision domain by leveraging the language modeling task, which is the core of natural language processing (NLP). Our main contribution is as follows: we have provided search space of recurrent neural networks on the text datasets and trained 14k architectures within it; we have conducted both intrinsic and extrinsic evaluation of the trained models using datasets for semantic relatedness and language understanding evaluation; finally, we have tested several NAS algorithms to demonstrate how the precomputed results can be utilized. We believe that our results have high potential of usage for both NAS and NLP communities.
\end{abstract}

\section{Introduction}
\label{sec:introduction}

NAS has matured as a recognized research field, numbering a few well-known difficulties, including the complicatedness of reproducibility and enormous computation costs. Reproducibility issues with NAS methods arise due to the variance in search spaces and experimental pipelines. Since NAS requires multiple runs of neural network training and black-box optimization, adherence to experimental protocol becomes of critical importance. Yang et al. \cite{yang2019evaluation} notice: most researchers do not strictly follow the experimental protocol or do not make enough ablation studies, leaving obscure reasons for the effectiveness of various methods. For instance, Li and Talwalkar \cite{li2019random} have been unable to exactly reproduce state-of-the-art (SOTA) NAS methods due to the lack of source code and data.  They have shown that random search with early stopping can achieve performance close to SOTA.

To address the reproducibility issue, a few recent works have proposed benchmarks. A NAS benchmark consists of numerous architectures trained and evaluated on a downstream task, with evaluation results stored for further utilization. The architectures and metrics thereby can be queried from the benchmark, so time- and energy-consuming training procedures can be avoided. Benchmarks play an important role in facilitating NAS research. However, the community still lacks the variety of benchmarks that allow fully reproducible NAS experiments, as Lindauer and Hutter pointed out \cite{lindauer2019best}. This justifies the need for the development of new and diverse NAS benchmarks.

Two largest benchmarks, NAS-Bench-101 \cite{ying2019bench} and NAS-Bench-201 \cite{Dong2020NAS-Bench-201:}, consist of convolutional and feedforward networks used for computer vision tasks. However, there are no NAS benchmarks that cover recurrent neural networks (RNN) or their modifications and natural language processing tasks. Thus, NAS applications to NLP have attracted fewer studies.

Many applications benefit from novel architectures and innovative design choices. For example, the quality of the core task in the natural language processing domain, language modeling \cite{ponte1998language,bengio2003neural}, was significantly improved by using Highway layers \cite{jozefowicz2016exploring,zilly2017recurrent} or tying input and output embeddings \cite{press2017using}. Residual connections \cite{he2016deep}  and dense connections  \cite{huang2017densely} were adopted from computer vision architectures and impacted on machine translation \cite{britz2017massive}, text classification \cite{wang2018densely}, and machine reading comprehension \cite{tay2018densely}. Such novel architectures can be mass-produced computationally employing NAS, which designs new architectures via solving an optimization problem. Efficient NAS methods are still to be researched, and in order to facilitate this process, we need to have reproducible benchmarks with well-defined frameworks. 

Creating a NAS benchmark is a challenging task; it includes several steps that have to be carefully designed and performed. First, the search space needs to be defined. Second, the datasets have to be selected. Finally, architectures from the search space need to be trained and evaluated according to both the objective function and to the downstream tasks' metrics.  

Our contributions are as follows:
\begin{enumerate}[leftmargin=20pt,noitemsep,topsep=0pt]
    \item We have presented {\bf the first RNN-derived NAS benchmark} designed for NLP tasks. Our benchmark has been derived from a {\bf novel search space} that comprises various RNN modifications,  including LSTM and GRU cells (Section \ref{sec:description}).
    \item We have trained {\bf over 14k architectures for the language modeling task}, and assessed the overall quality of each architecture in terms of language modeling and two additional setups. We have conducted an intrinsic evaluation of learned static word embeddings by applying word similarity tests and an extrinsic evaluation by measuring the performance in downstream tasks, such as GLUE benchmarks (Section \ref{sec:analysis}).
    \item We have introduced a framework for benchmarking and conducted a thorough comparison of different NAS algorithms within it (Section \ref{sec:benchmark}). In particular, the framework provides a convenient proper way to simulate and measure the training wall time of a NAS process.
    \item We have {\bf released all learned architectures}, which allows architecture comparison in many aspects. A representative subset of architectures is provided with {\bf metrics for language modeling and downstream tasks}.
\end{enumerate}

The source code and links to precomputed files are available in this repository:

\vspace{-6pt}
\url{https://github.com/fmsnew/nas-bench-nlp-release}

\vspace{-9pt}






\section{Related work}

\textbf{NAS benchmarks}

Two earliest benchmarks were released during 2019: NAS-Bench-101 \cite{ying2019bench} and NAS-HPO-Bench \cite{klein2019tabular}. NAS-Bench-101 \cite{ying2019bench} comprises 423k unique convolutional architectures trained on CIFAR-10 image dataset. NAS-HPO-Bench \cite{klein2019tabular} latter comprises joint hyperparameters and architecture search space with a total size of 62k unique feedforward architectures and hyperparameters configurations evaluated on four regression datasets. Both benchmarks entirely cover their search spaces within the defined constraints. However, each network cell is of very low complexity, that limits sophisticated feature engineering because instead of using e.g., graph neural networks, one can merely encode architectures with their graphs adjacency matrices.   

Two later projects elaborated on the NAS-Bench-101 benchmark. NAS-Bench-201 \cite{Dong2020NAS-Bench-201:} by Dong and Yang proposed a framework and ran ten popular NAS algorithms within it. This benchmark contains 15.6k architectures trained on three image classification datasets, however, cell sizes are even smaller: they have only five nodes compared to the maximum of seven nodes in NAS-Bench-101. 
Zela et al. \cite{zela2020bench} built a one-shot NAS framework on the top of NAS-Bench-101 that can reuse the underlying computations. Several one-shot NAS methods were adjusted to query approximate instances from NAS-Bench-101. However, this extension works only with sub-spaces of the original NAS-Bench-101 search space.

\textbf{NAS for RNN}

A few previous studies have attempted to optimize existing conventional recurrent cells and concluded that they are not necessarily optimal, moreover, the significance of their components is unclear \cite{greff2016lstm,jozefowicz2015empirical}.

Greff et al. \cite{greff2016lstm} conducted a study of eight modifications of LSTM architecture (Long-Short Term Memory \cite{hochreiter1997long}) on several tasks. They concluded that the ordinary LSTM performs reasonably well on all datasets, and no modification improves its performance significantly. However, some of the modifications also look promising due to the lower number of parameters and similar performance.

Jozefowich et al \cite{jozefowicz2015empirical} tried to find an architecture that outperforms LSTM using evolutionary approach applied to GRU (Gated Recurrent Unit \cite{cho2014properties}) and LSTM. They evaluated over ten thousand modifications and did not find any  that would consistently outperform LSTM on various datasets, although a few instances had superior performance on particular datasets. The authors concluded that architectures that dramatically outperform LSTM could not be easily found in the local area around the vanilla configuration. 

There are also the works that introduce algorithms to increase efficiency of NAS, in particular they include application of those algorithms for designing RNN cells from scratch \cite{zoph2016neural,liu2018darts,pham2018efficient}. The experimental results have shown that these methods are capable of obtaining competitive results with SOTA.

\section{Description}
\label{sec:description}

The benchmark in this work has the following components in its foundation: datasets, search space, training and evaluation protocols. In this section we describe all of them.

\begin{wraptable}{r}{5.5cm}
    \caption{Statistics of language modeling datasets.}
    \label{tbl:datasets_stats}
    \centering
    \begin{tabular}{ccc}
        \toprule
        Dataset & PTB & WikiText-2 \\
        \midrule
        Tokens & 1.086M & 2.552M \\
        Vocab size & 10000 & 33278\\
        OoV rate\footnotemark & 0.049 & 0.032 \\
        \bottomrule
    \end{tabular}
    \vspace{-15pt}
\end{wraptable}
\footnotetext{Out of Vocabulary (OoV) rate --- share of tokens outside of the model's vocabulary; all OoV tokens are replaced with a service token <unknown>.}

    
\textbf{Datasets}

We use Penn Tree Bank (PTB) \cite{mikolov2010recurrent} to train a sample of networks from the search space. We also use the WikiText-2 \cite{merity2016pointer} dataset to train a stratified subsample of networks based on their performance on the PTB dataset.  The second dataset is larger and is more realistic since it preserves the letter case, punctuation, and numbers.  Statistics of these datasets are shown in Table \ref{tbl:datasets_stats}.

\textbf{Macro-level of the search space (AWD-LSTM)}

\begin{figure}[ht]
\centering
\vspace{-10pt}
\includegraphics[width=0.9\textwidth]{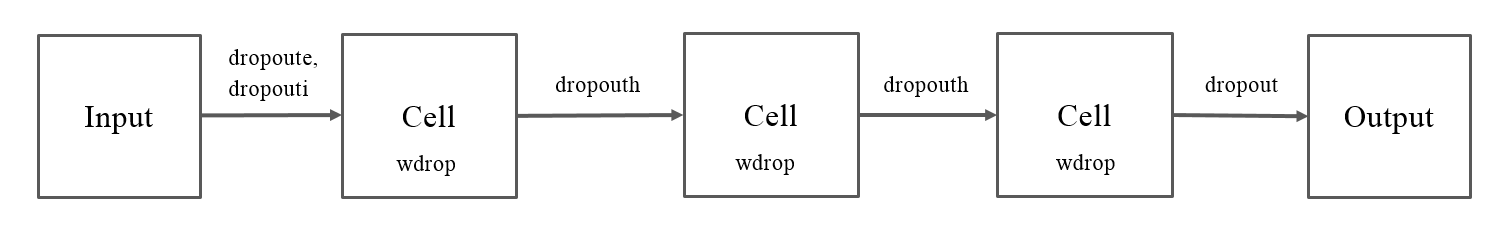}
\setlength{\belowcaptionskip}{-10pt}
\caption{AWD-LSTM macro-level. See definitions of dropouts and other parameters of the architecture in the original repository \cite{awd_lstm_repo}.}
\label{fig:awd-lstm}
\end{figure}

The macro-level of each model and the training procedure are borrowed from AWD-LSTM \cite{merity2017regularizing} as it has a relatively simple structure and has comparable to SOTA performance. The network consists of three stacked cells with weightdrop regularizations in each and locked dropouts between them and input/output layers, as well as a dropout applied to input embedding (see Figure \ref{fig:awd-lstm}).

\textbf{Micro-level of the search space (recurrent cells)}

\begin{figure}[ht]
\centering
\begin{subfigure}{0.32\textwidth}
    \centering
    \includegraphics[width=0.7\textwidth]{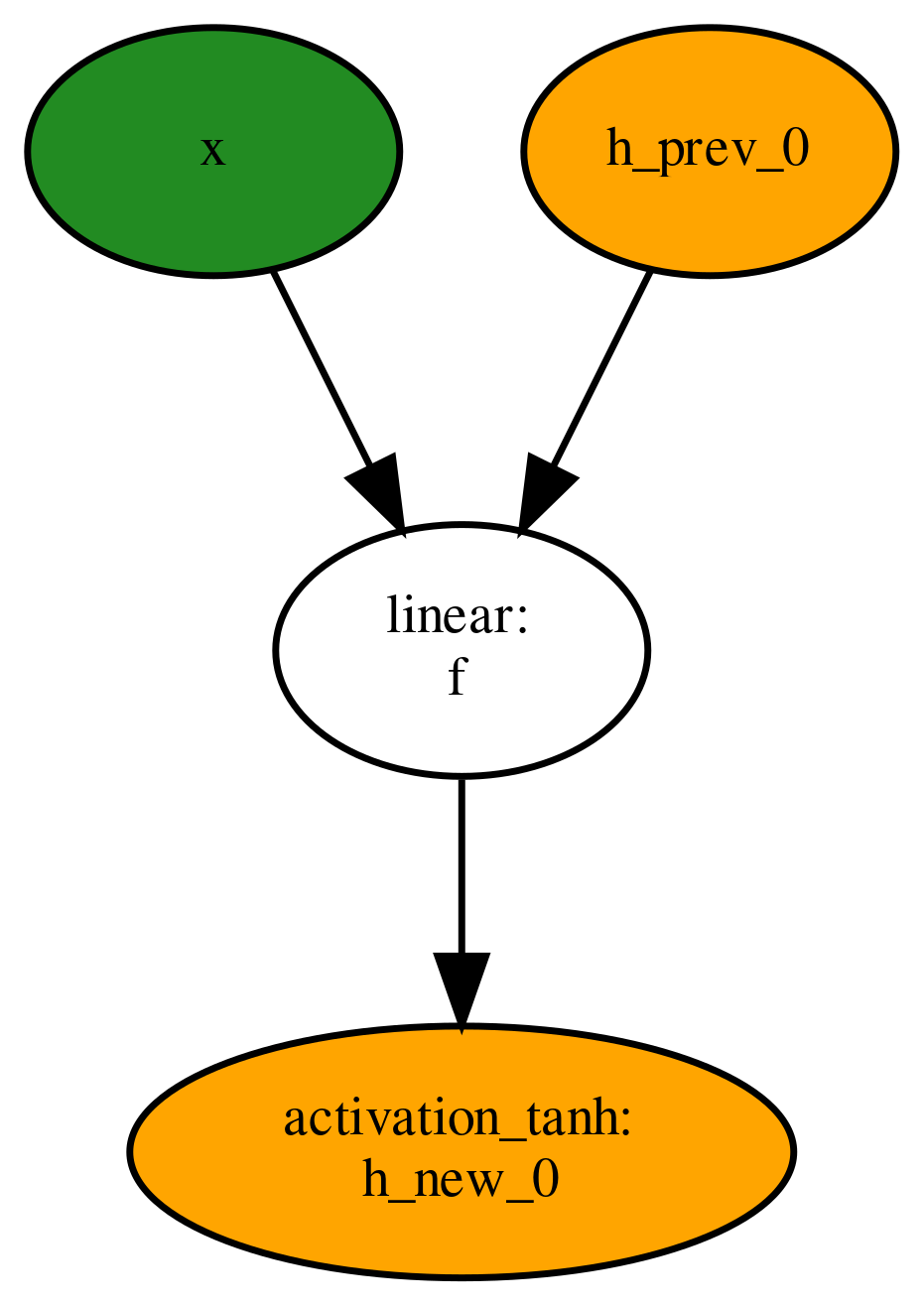}
    \caption{Simple RNN Cell}
    \label{fig:rnn_cell}
\end{subfigure}
\begin{subfigure}{0.32\textwidth}
    \centering
    \includegraphics[width=1.0\textwidth]{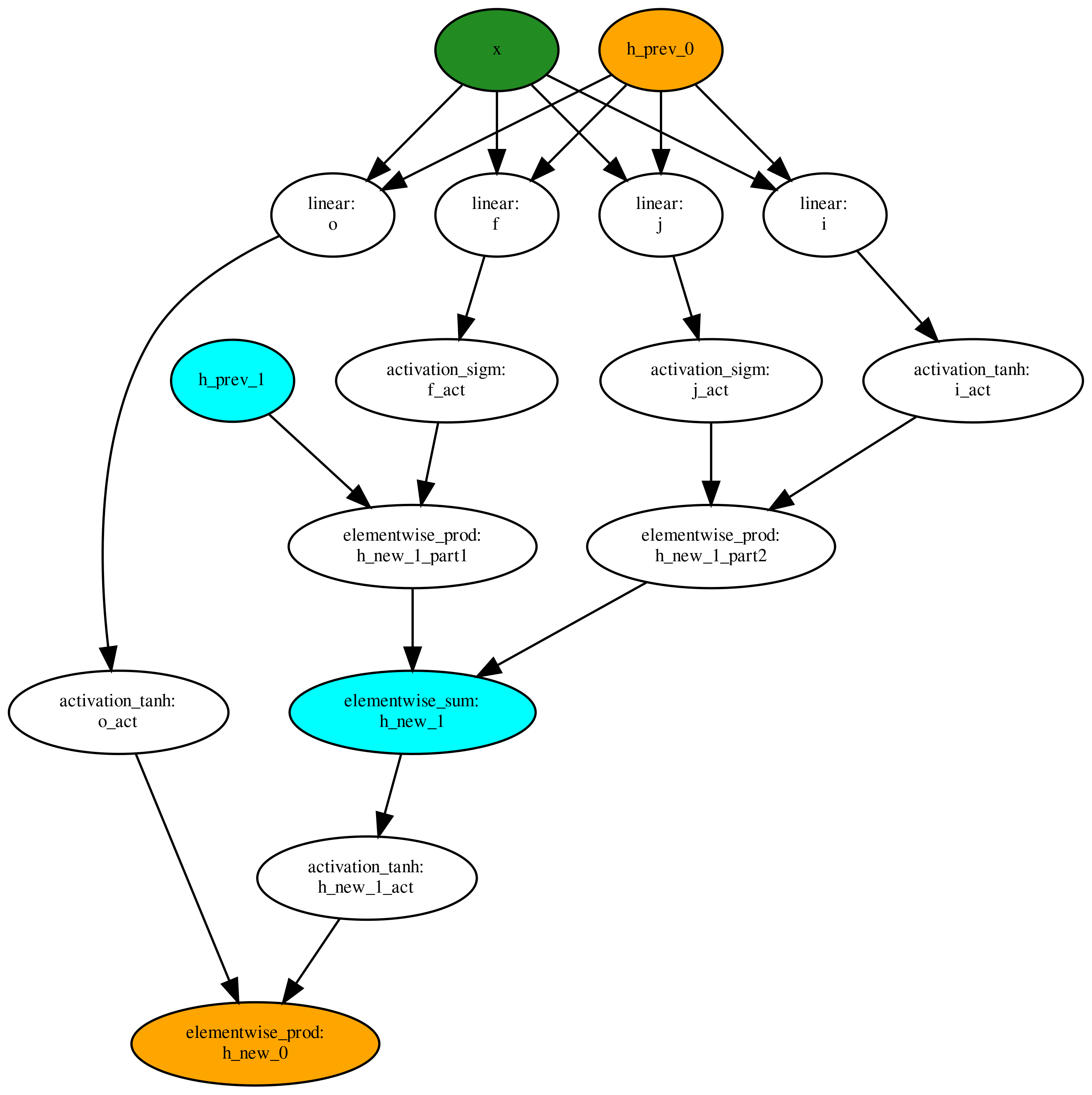}
    \caption{LSTM Cell}
    \label{fig:lstm_cell}
\end{subfigure}
\begin{subfigure}{0.32\textwidth}
    \centering
    \includegraphics[width=0.6\textwidth]{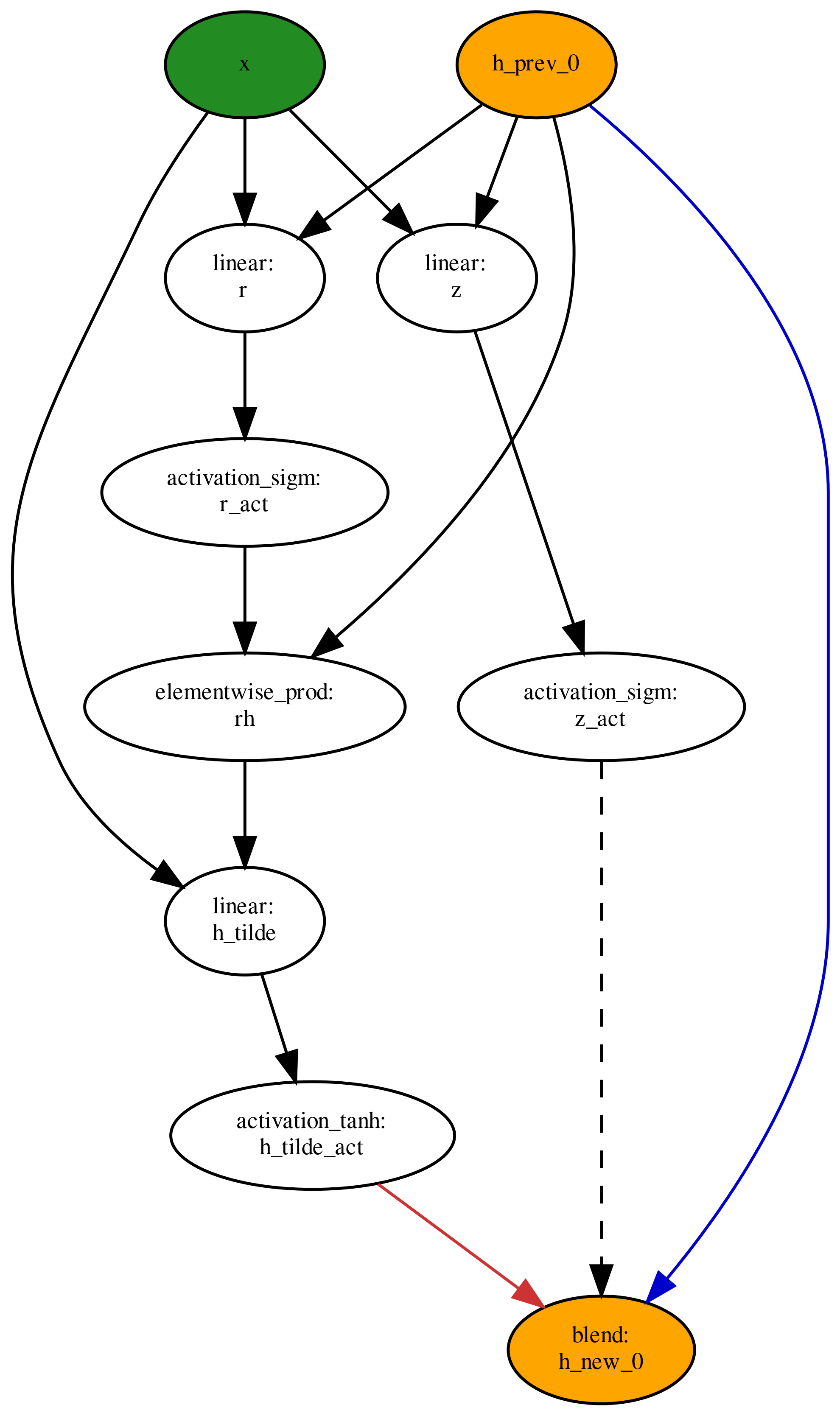}
    \caption{GRU Cell}
    \label{fig:gru_cells}
\end{subfigure}
\setlength{\belowcaptionskip}{-5pt}
\caption{Examples of conventional RNN cells. Colors of nodes highlight the corresponding previous and new hidden states, green color also highlights the input vector. Black dashed, blue and red edges indicate blending arguments $z$, $x$ and $y$ respectively.}
\label{fig:rnn_cells}
\end{figure}

We define a search space for cells (micro-level of models) to include all conventional recurrent cells as particular instances (see examples in Figure \ref{fig:rnn_cells}). Cell computations are encoded as attributed graphs: each node is associated with an operation, and edges encode its inputs. The following operations are available:
\begin{itemize}[leftmargin=20pt,noitemsep,topsep=0pt]
    \item[-] Linear: $f(x_1, .., x_n) = W_1 x_1 + .. + W_n x_n + b$,
    \item[-] Blending (element wise): $f(z, x, y) = z \odot x + (1 - z) \odot y$,
    \item[-] Element wise product and sum,
    \item[-] Activations: Tanh, Sigmoid, and LeakyReLU.
\end{itemize}
We impose some constraints on possible instances: number of nodes $\leq 24$, number of hidden states $\leq 3$, number of linear input vectors $n \leq 3$.

\textbf{Generation procedure}

We used the following graph generation procedure: initial nodes correspond to the input vector and hidden states; at each step, a node is added, an operation associated to the node is randomly chosen and depending on the operation, connections are made with previous nodes; after all, nodes are added, new hidden states are randomly selected among them; next, redundant nodes, that do not lead to the new hidden states in the computational graph, are removed; finally, the architecture is accepted if the input vector and hidden state nodes are in the graph (not redundant), and no hidden state node is directly connected to the new hidden states in order to avoid numeric explosions.

In addition, we manually added three architectures to the generated sample as baselines: RNN (Simple RNN Cell), LSTM, and GRU (see Figure \ref{fig:rnn_cells}).

\textbf{Training process}

We generated 14322 architectures for training on the PTB dataset, 4114 of them were trained three times with different random seeds, the others were trained once; also, 289 out of them were trained on WikiText-2 based on stratified perplexities for PTB.

\begin{figure}[ht]
\centering
\begin{subfigure}[t]{0.46\textwidth}
    \centering
    \includegraphics[width=1.0\textwidth,valign=t]{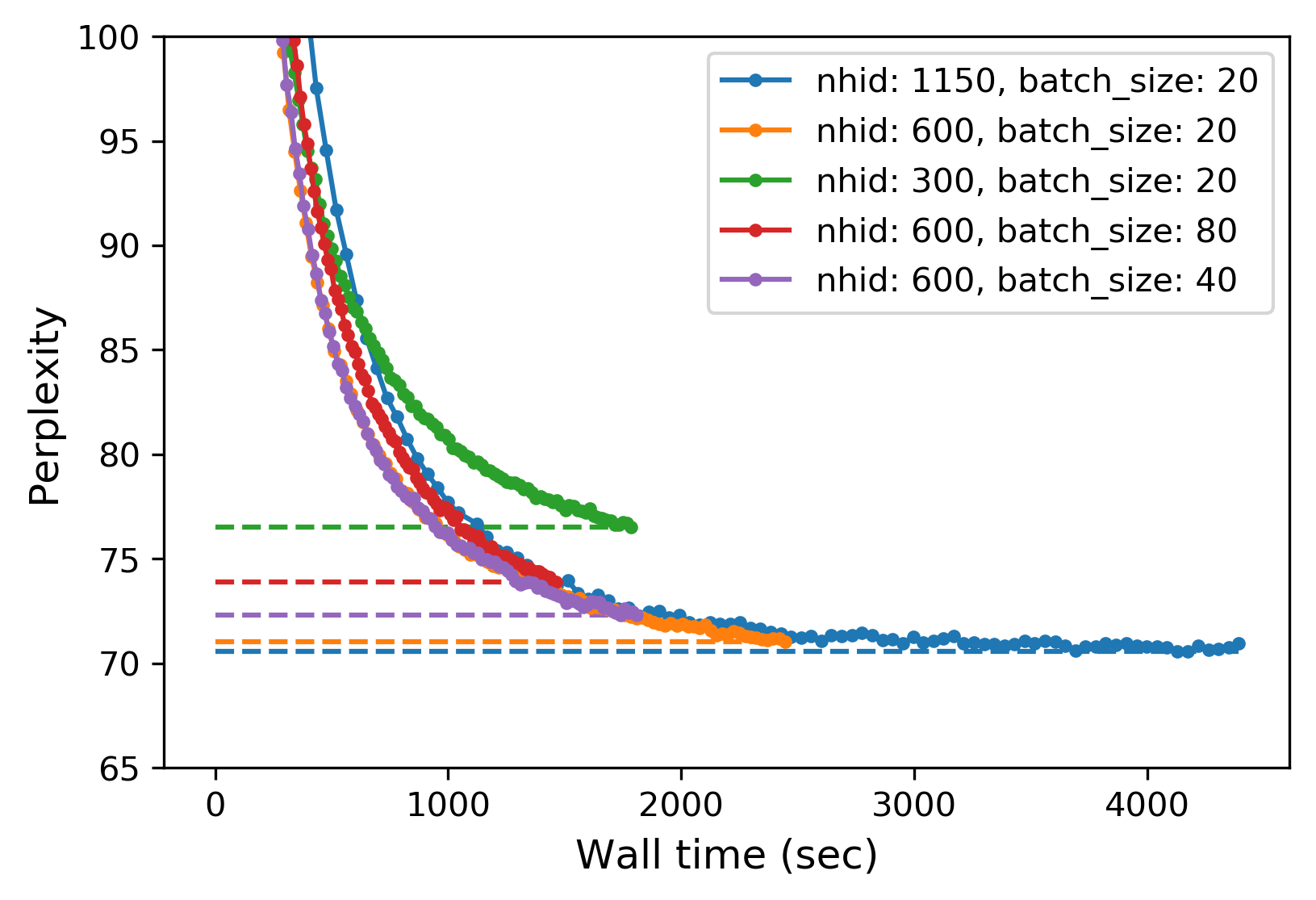}
    \vphantom{\includegraphics[width=0.07\textwidth,valign=t]{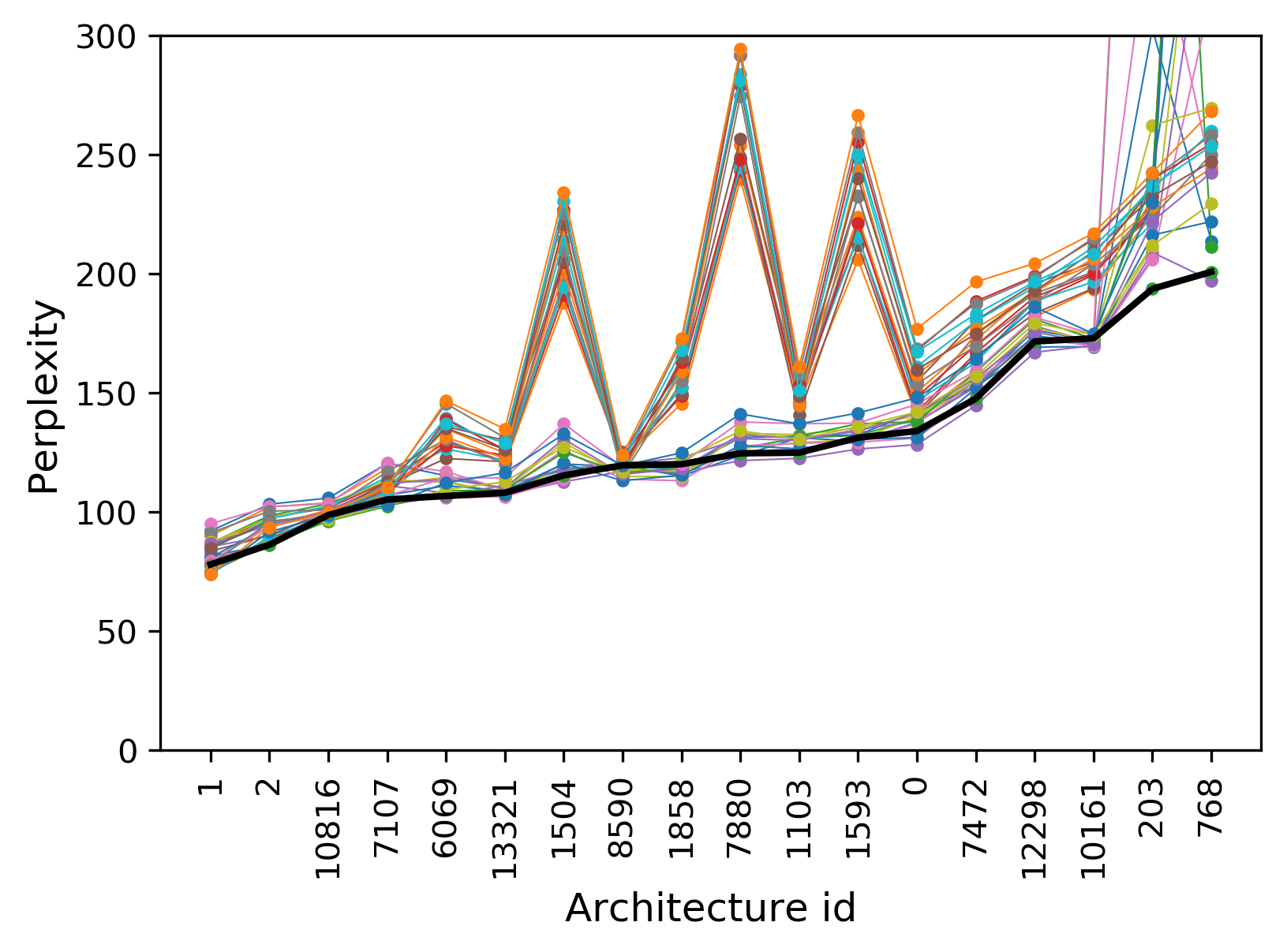}} 
    \caption{Hidden states size (nhid) and batch size for AWD-LSTM.}
    \label{fig:hp_nhid_batch}
\end{subfigure}
\hspace{10pt}
\begin{subfigure}[t]{0.46\textwidth}
    \centering
    \includegraphics[width=1.0\textwidth,valign=t]{assets/hyper_selection.png}
    \caption{Dropouts. Curves correspond to various dropout configuration; black curve corresponds to the best one on average.}
    \label{fig:hp_dropouts}
\end{subfigure}
\caption{Hyper-parameters selection.}
\label{fig:hp_selection}
\end{figure}

First, we found a trade-off between training time and validation performance on PTB after 50 iterations by varying sizes of the hidden states and batch size for AWD-LSTM (Figure \ref{fig:hp_nhid_batch}). The chosen pair was $nhid=600$ and $batch\_size=20$ because such a network almost converges to the same perplexity as the original AWD-LSTM, where $nhid=1150$, but within a half of the original training time.  
Then we selected a random subset of architectures and performed a grid search of dropout values for each of them on PTB. Figure \ref{fig:hp_dropouts} shows the performance of all configurations for the selected architectures. The following configuration showed the best validation perplexity on average: $dropout=0.1$, $dropouth=0.25$, $dropouti=0.4$, $dropoute=0.0$, and $wdrop=0.1$. We used that configuration and fixed other training settings for all architectures on both datasets.

\textbf{Evaluation}

We logged the following models metrics for each epoch: wall time and train/validation/test log perplexity $\textup{log\_PPL}(p) = -\sum_x p(x) \log p(x)$, where $p$ is a discrete probability distribution of words. In addition, the total number of trainable parameters and their final values were stored to evaluate architectures on downstream tasks. 

\textbf{Hardware}

We use the following hardware specifications to precompute architectures:
HPC cluster Zhores \cite{zacharov2019zhores} with Tesla V100-SXM2 with 16Gb of memory on each GPU.
\section{Analysis}
\label{sec:analysis}

\subsection{Search space evaluation}
The complete search space is extremely large; for example, there are approximately $10^{53}$ connected non-attributed graphs with $\leq 24$ nodes and $\leq 3$ input edges. 

Some architectures from the generated sample (around 23\%) turned out to suffer from numerical explosions that occurred when, for example, there were no activations between corresponding hidden states.

\begin{figure}[ht]
\centering
\begin{subfigure}{0.35\textwidth}
    \centering
    \includegraphics[width=1.0\textwidth]{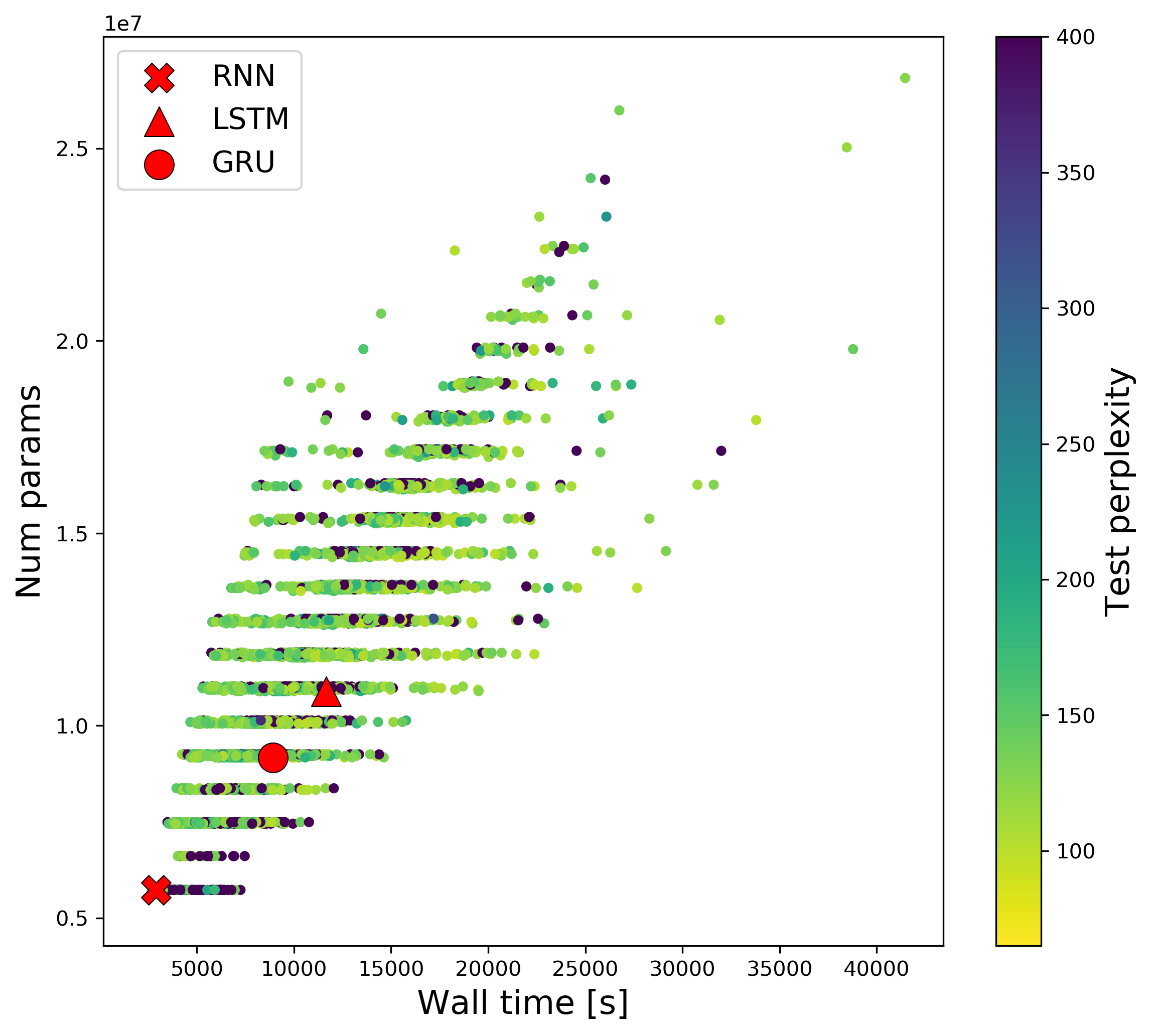}
    \caption{Joint distribution of metrics.}
    \label{fig:arch_metrics_joint}
\end{subfigure}
\hspace{15pt}
\begin{subfigure}{0.46\textwidth}
    \centering
    \includegraphics[width=1.0\textwidth]{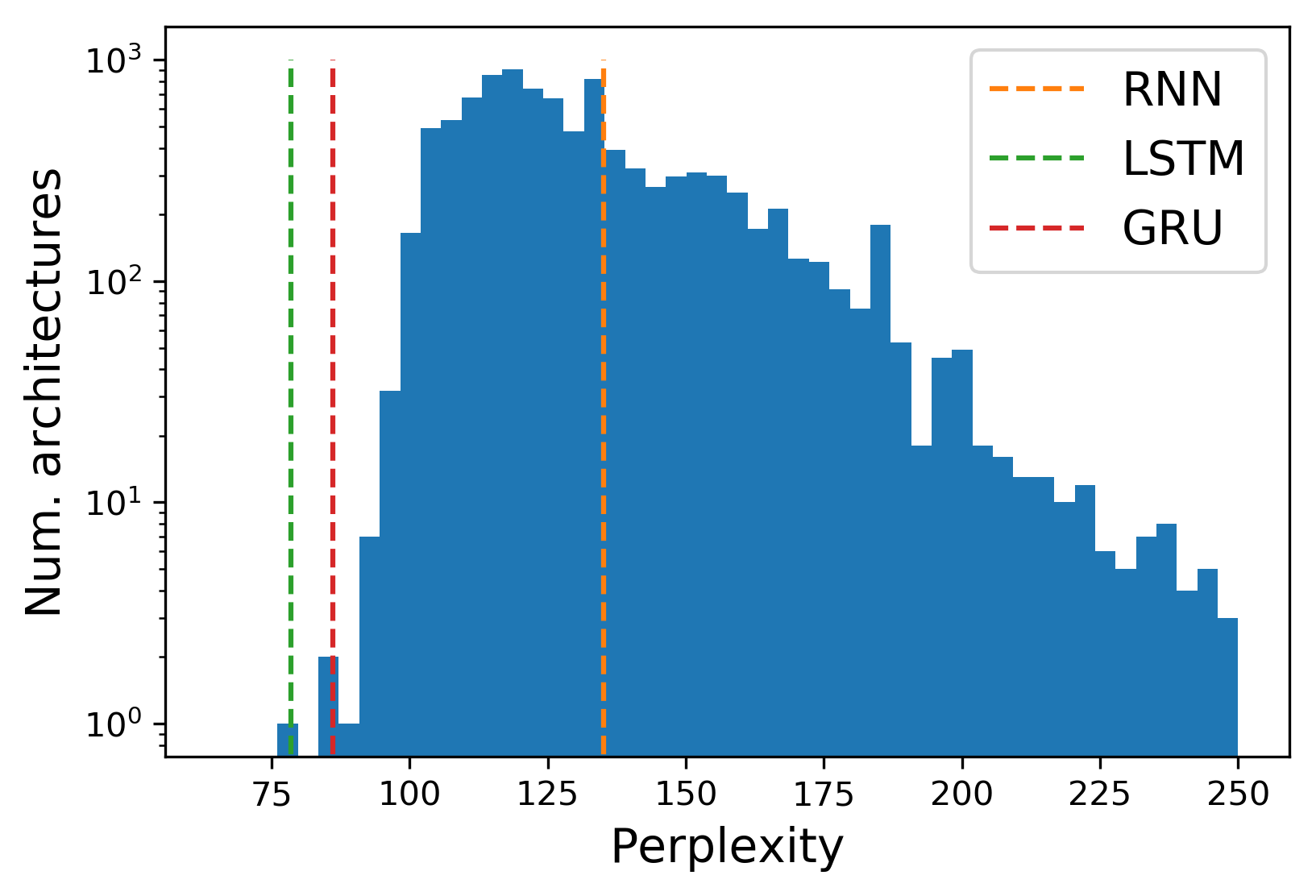}
    \caption{Best test perplexity distribution.}
    \label{fig:arch_ppl_distrib}
\end{subfigure}
\caption{Architectures metrics on PTB.}
\label{fig:arch_metrics}
\end{figure}

Figure \ref{fig:arch_metrics_joint} shows the relationship among three metrics: number of parameters, training wall time, and test perplexity. According to the plot, there is no clear correlation between the test perplexity and the number of parameters or training wall time; LSTM and GRU also look like typical representatives of the generated sample. However, based on figure \ref{fig:arch_ppl_distrib}, which shows the distribution of the best test perplexity achieved by each architecture, LSTM and GRU architectures have the top performance in terms of the test perplexity, whereas RNN looks average.

\begin{figure}[ht]
\centering
\includegraphics[width=0.99\textwidth]{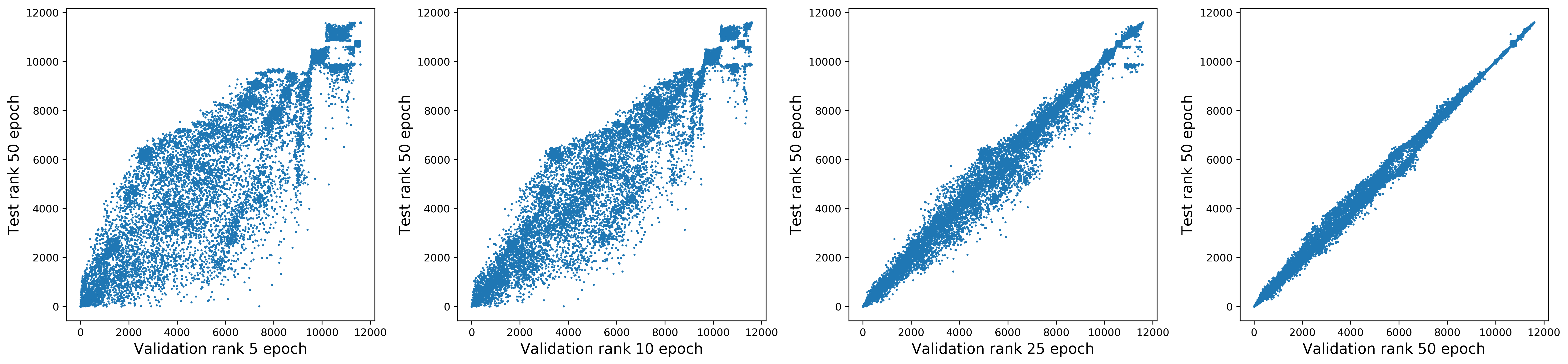}
\caption{The ranking of architectures perplexity for different epochs on test and validation sets of PTB: lower rank corresponds to lower perplexity.}
\label{fig:dynamic_ranking}
\end{figure}

We have investigated how the validation perplexity at different training stages correlates with the final testing perplexity. Figure \ref{fig:dynamic_ranking} shows the ranking of architectures based on their final testing perplexities w.r.t. ranking obtained with validation perplexities at 5, 10, 25, and 50 epochs.

\begin{figure}
\centering
\begin{minipage}{.24\linewidth}
  \centering
    \includegraphics[width=1.0\textwidth]{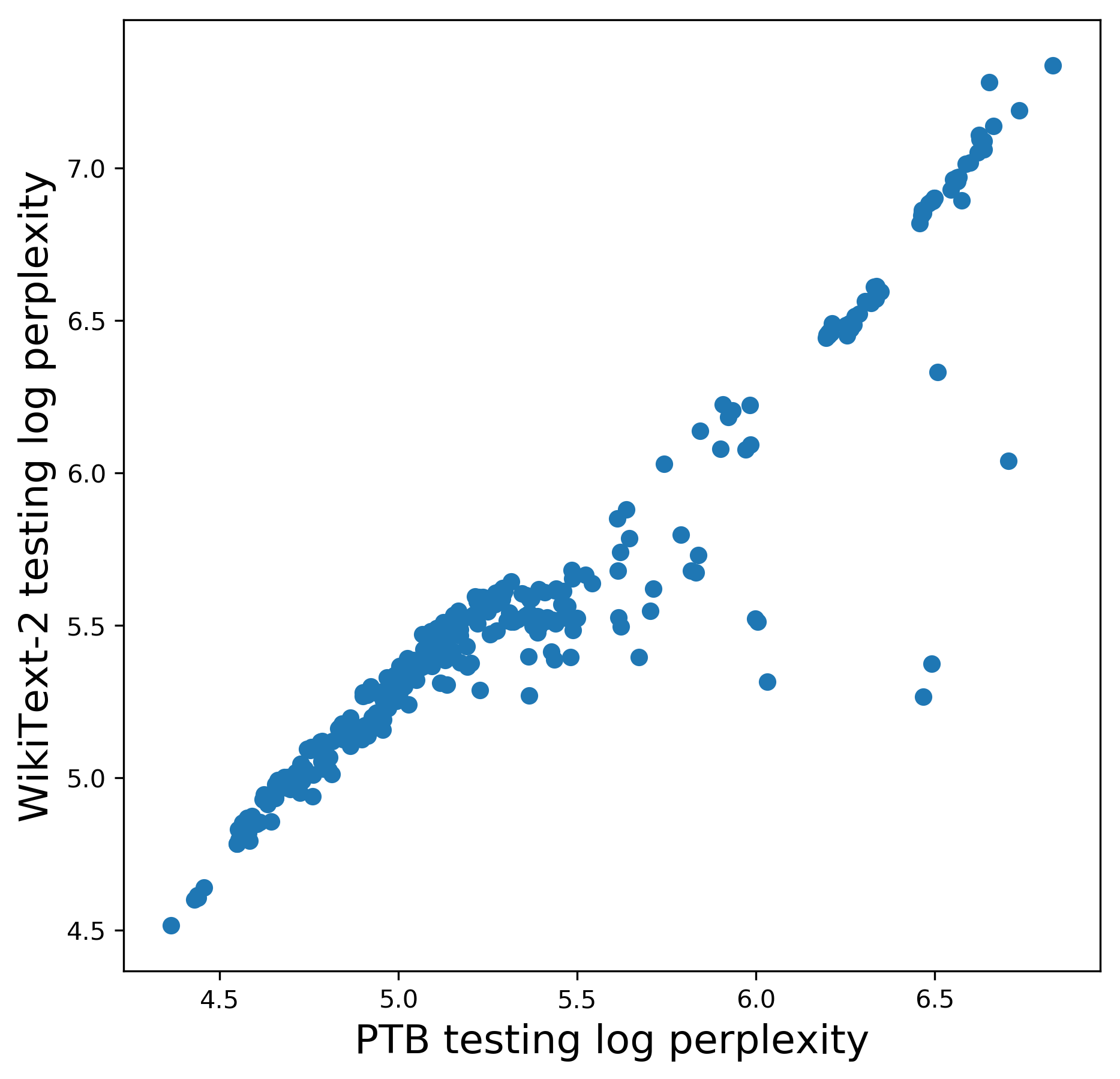}
    \captionof{figure}{Correlation of architectures' test log perplexities on PTB and WikiText-2.}
    \label{fig:arch_transfer}
\end{minipage}%
\hspace{20pt}
\begin{minipage}{.35\linewidth}
  \centering
    \includegraphics[width=1.\textwidth]{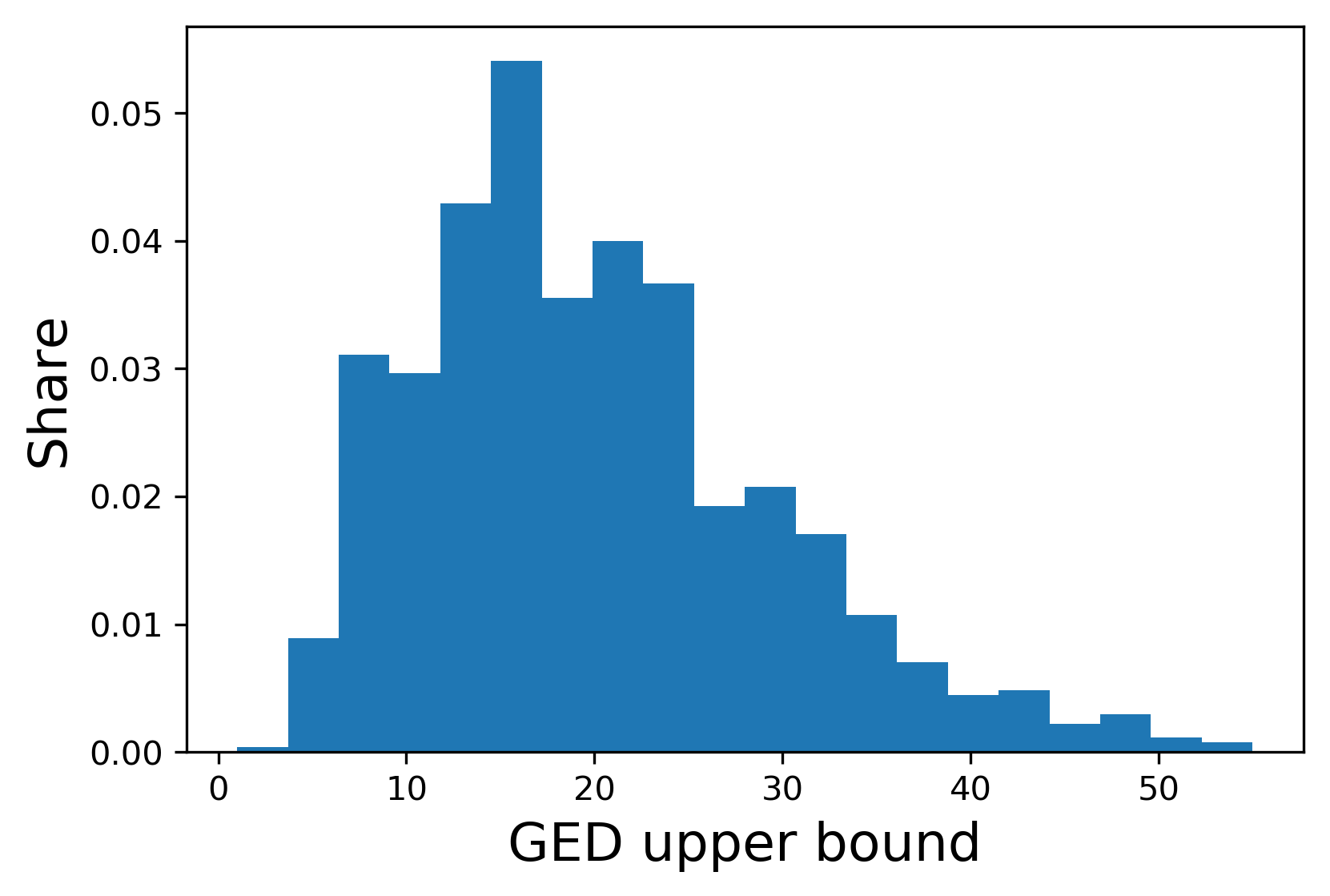}
    \captionof{figure}{Histogram of upper bounds of graph edit distances between 1000 random pairs of architectures.}
    \label{fig:ged}
\end{minipage}
\hspace{20pt}
\begin{minipage}{.24\linewidth}
  \centering
    \includegraphics[width=1.0\textwidth]{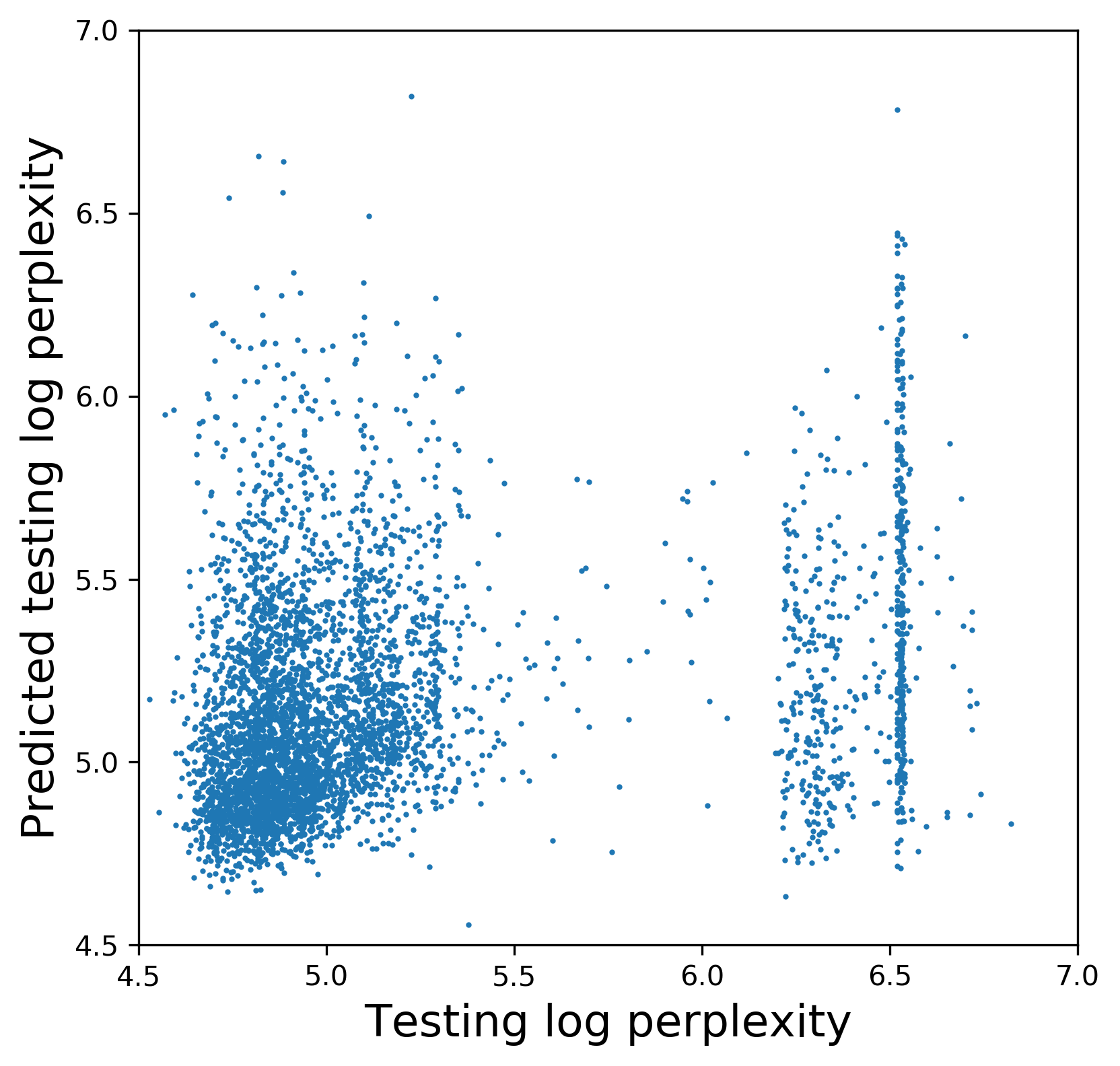}
    \captionof{figure}{True vs predicted final testing log perplexity based on graph2vec features.}
    \label{fig:pred_loss}
\end{minipage}%

\end{figure}

Figure \ref{fig:arch_transfer} shows the correlation of architectures performance on PTB and WikiText-2. The plot suggests good transfer properties of NAS for RNN, that is, architectures that perform well on one dataset, will also perform well on another.

We have investigated the sparsity of the generated sample of architectures. Figure \ref{fig:ged} shows the histogram of upper-bounds on Graph Edit Distances (GED) between 1000 random pairs of architectures, which also take into consideration the difference in operations associated with each node. To calculate these values, we used consecutive GED approximations \cite{abu2015exact} (implemented in NetworkX package \cite{hagberg2008exploring}) for the limited time. 

\begin{wraptable}{r}{7.5cm}
    \vspace{-10pt}
    \caption{Detailed comparison of top performing architectures and ordinary RNN on PTB dataset.}
    \label{tbl:selected_performance}
    \centering
    \begin{tabular}{ccc}
        \toprule
        Architecture & Num. params & Test perplexity \\
        \midrule
        LSTM & 10.9M & 78.5 \\
        Top-2 & 9.2M & 84.7\\
        GRU & 9.2M & 86.1 \\
        Top-4 & 11.0M & 90.6 \\
        \midrule
        RNN & 5.73M & 135.1 \\
        \bottomrule
    \end{tabular}
    \vspace{-20pt}
\end{wraptable}
Table \ref{tbl:selected_performance} compares the top architectures in more detail. LSTM has the best performance, GRU also achieves top-3 performance, however, the generated sample also contains a few competitive examples that are substantially different from LSTM and GRU. Their architectures are provided in Appendix \ref{sec:appendix_top_archs}, Figure \ref{fig:best_archs}.

\textbf{Parameterization of architectures}

We used graph2vec \cite{narayanan2017graph2vec} method to create the characteristic features for architectures, which can be useful for some NAS methods like Bayesian Optimization \cite{frazier2018tutorial}. With this method, architectures were embedded in 10- and 50-dimensional spaces. In order to verify that the obtained features were sensible, we used them to 1) classify the flawed architectures, the ones that experienced problems during the training; and 2) predict the final testing log perplexity. We split corresponding datasets into equal training and testing parts and used XGBoost \cite{chen2016xgboost} for both problems. ROC AUC metric for the classification (1) was 0.98, whereas $R^2$-score for the regression task (2) was 0.012 (see also Figure \ref{fig:pred_loss}). We also trained and evaluated task (2) on architectures with log perplexity $\leq 6$, and obtained $R^2$-score 0.24.

\subsection{Word embedding evaluation}
Approaches to language model evaluation fall into two major categories: intrinsic and extrinsic evaluation \cite{schnabel2015evaluation}. Intrinsic evaluation tests the semantic relationships between words \cite{baroni2014don,mikolov2013distributed}, and, thus leverages word embeddings only. The whole language model is not involved in the evaluation. The first layer solely, i.e., static (fixed) word embeddings, is subjected to evaluation. 
Extrinsic evaluation treats the language model as dynamic (contextualized) word embeddings and feeds them as an input layer to a downstream task. It measures the performance metrics for the task to assess the quality of the language model.
To fully assess trained architectures, we subject them to intrinsic and extrinsic evaluation, following the best practices from the NLP community. For the sake of time, we did not assess all architectures. Instead, we used a stratified sample of architectures according to the perplexity values.

\newpage
\textbf{Intrinsic evaluation}

We picked two benchmark datasets to evaluate semantic relationships between word pairs: WordSimilarity-353 \cite{finkelstein2001placing}, consisting of similatiry judgments for 353 word pairs on a scale of 0 to 10. For example, word pair ``book'' and ``paper'' are scored with 7.5. SimLex-999   \cite{hill2015simlex} is a larger and more challenging benchmark, consisting of similarity judgments for  999 word pairs. To evaluate word embeddings for each word pair, cosine similarity between word vectors is computed. The resulting similarity values are correlated with the judgments using Spearman's and Pearson's correlation coefficients. 
We used the benchmarks distributed by the gensim framework\footnote{\url{https://radimrehurek.com/gensim/}}.
To ensure a fair comparison, we train two word2vec models (in particular, Skip-gram with negative sampling, SGNS)  \cite{mikolov2013distributed} on PTB and WikiText-2 independently and use them as baselines for both intrinsic and extrinsic evaluation. 

\begin{figure}[!ht]

\begin{subfigure}[b]{.46\linewidth} 
\includegraphics[width=\textwidth]{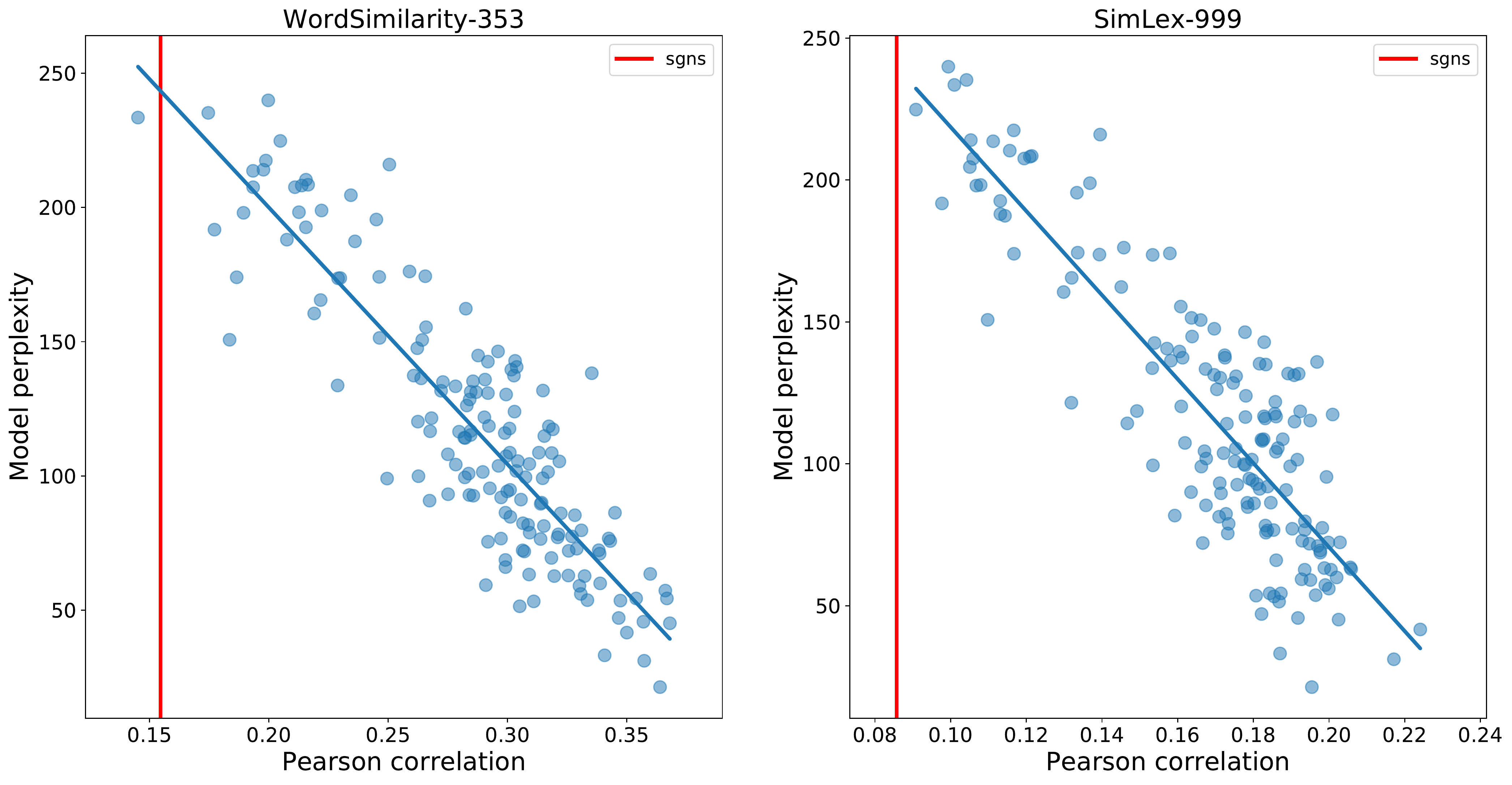}
\caption{Architectures trained on PTB}
\end{subfigure}
\begin{subfigure}[b]{.46\linewidth} 

\includegraphics[width=\textwidth]{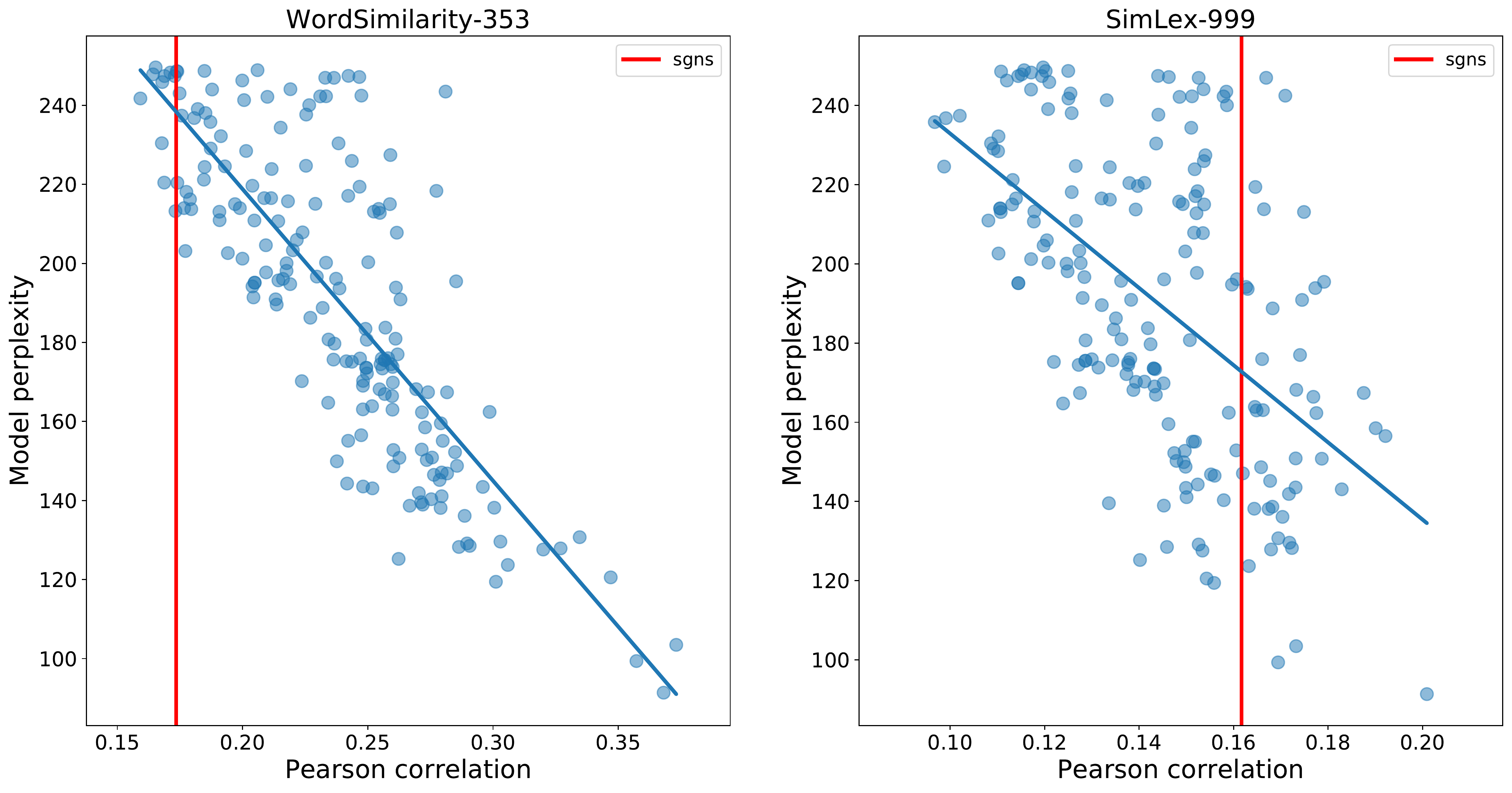}
\caption{Architectures trained on WikiText-2}
\end{subfigure}

\setlength{\belowcaptionskip}{-10pt}
\caption{OX: performance of 150 random architectures on WordSimilarity-353 and SimLex-999 (measured by Pearson correlation coefficient), OY: model perplexety, red line: SGNS performance on WordSimilarity-353 and SimLex-999 (measured by Pearson correlation coefficient).}
\label{fig:static_we_eval}
\end{figure}

Figure ~\ref{fig:static_we_eval} shows the results of intrinsic evaluation. When trained on PTB, most of our architectures overcame SGNS by a large margin, judging from the evaluation on both benchmarks. However, this evaluation is overestimated as the intersection of the PTB vocabulary and both benchmarks are rather small. Similar patterns are observed when both SGNS and our architectures are trained on WikiText-2 and compared on WordSim-353. However, as SimLex-999 is more challenging than WordSim-353, only a half of the architectures manages to beat the SGNS baseline. As expected, in all settings, the lower the perplexity is, the higher the performance of the models is.

\textbf{Extrinsic evaluation}

We used the General Language Understanding Evaluation (GLUE) benchmark \cite{wang2018glue}, a collection of ten diverse tasks aimed at the evaluation of language models performance. The GLUE score is computed as an average of all performance measures for ten tasks, multiplied by 100.
  
We follow the GLUE evaluation pipeline \cite{wang2018glue}: for each task, our architectures encode the input sentences to the vectors, which are passed further to a classifier. We adjust the Jiant toolkit \cite{wang2019jiant} to process our architectures. 

Finally, we evaluate two baselines: 1) the average bag-of-words using SGNS embeddings and 2) an LSTM encoder with SGNS embeddings in the same setting. When trained on PTB, only a few architectures perform better than simple bag-of-words baselines. None outperforms the LSTM baseline. Due to the small size of these models vocabulary, we do not observe any dependence between the perplexity values and the GLUE score. The architectures that pose a larger vocabulary, trained on WikiText-2, cope with GLUE tasks much better, and almost 20\%  of them beat both baselines, achieving mean GLUE at the level of 42.

To conclude,  our results show that the architectures reach baseline performance and even exceed it in several NLP benchmarks if compared to strong baselines trained under the same conditions. However, these architectures do not achieve the same performance as the recent Transformer-based models, such as BERT \cite{kenton2019bert}, T5\cite{raffel2019exploring}, or ELECTRA \cite{clark2019electra}.

\section{NAS Benchmark}
\label{sec:benchmark}

We prepared an environment that simulates NAS processes and does the proper measurements of metrics. The environment can perform the following tasks:
\begin{itemize}[leftmargin=20pt,noitemsep,topsep=0pt]
    \item train an architecture for the specified number of epochs (the environment automatically simulates checkpoints and continuation of the training process);
    \item return architecture metrics at the specific training epoch (the architecture must be trained until the requested epoch); 
    \item return total simulated wall time;
    \item return the testing log perplexity of the best configuration (architecture and epoch) based on validation perplexity.
\end{itemize}

For benchmarking NAS algorithms on our search space, we tested the following methods within the environment: 
\begin{itemize}[leftmargin=20pt,noitemsep,topsep=0pt]
    \item Random Search in two modes: low-fidelity mode (RS 10E), which trains 5X networks for 10 epochs, and high-fidelity mode, which trains 1X networks for 50 epochs (RS 50E);
    \item Hyperbands (HB) is a feature-agnostic multi-fidelity method \cite{li2017hyperband};
    \item Bayesian Optimization using 10-dimensional (BO 10D) and 50-dimensional (BO 50D) graph2vec features (Sec. \ref{sec:analysis}, parameterization of architectures) and a bagged \cite{breiman1996bagging} XGBoost regressor to estimate uncertainty of predictions;
    \item Regularized Evolution (RE) \cite{real2019regularized} using graph2vec features;
    \item Hyperopt with the Tree-structured Parzen Estimator (TPE); \cite{bergstra2013making};
    \item SMAC \cite{hutter2011sequential} using 10-dimensional graph2vec features.
\end{itemize}

\begin{wrapfigure}{r}{6.5cm}
\centering
\vspace{-10pt}
\includegraphics[width=0.4\textwidth]{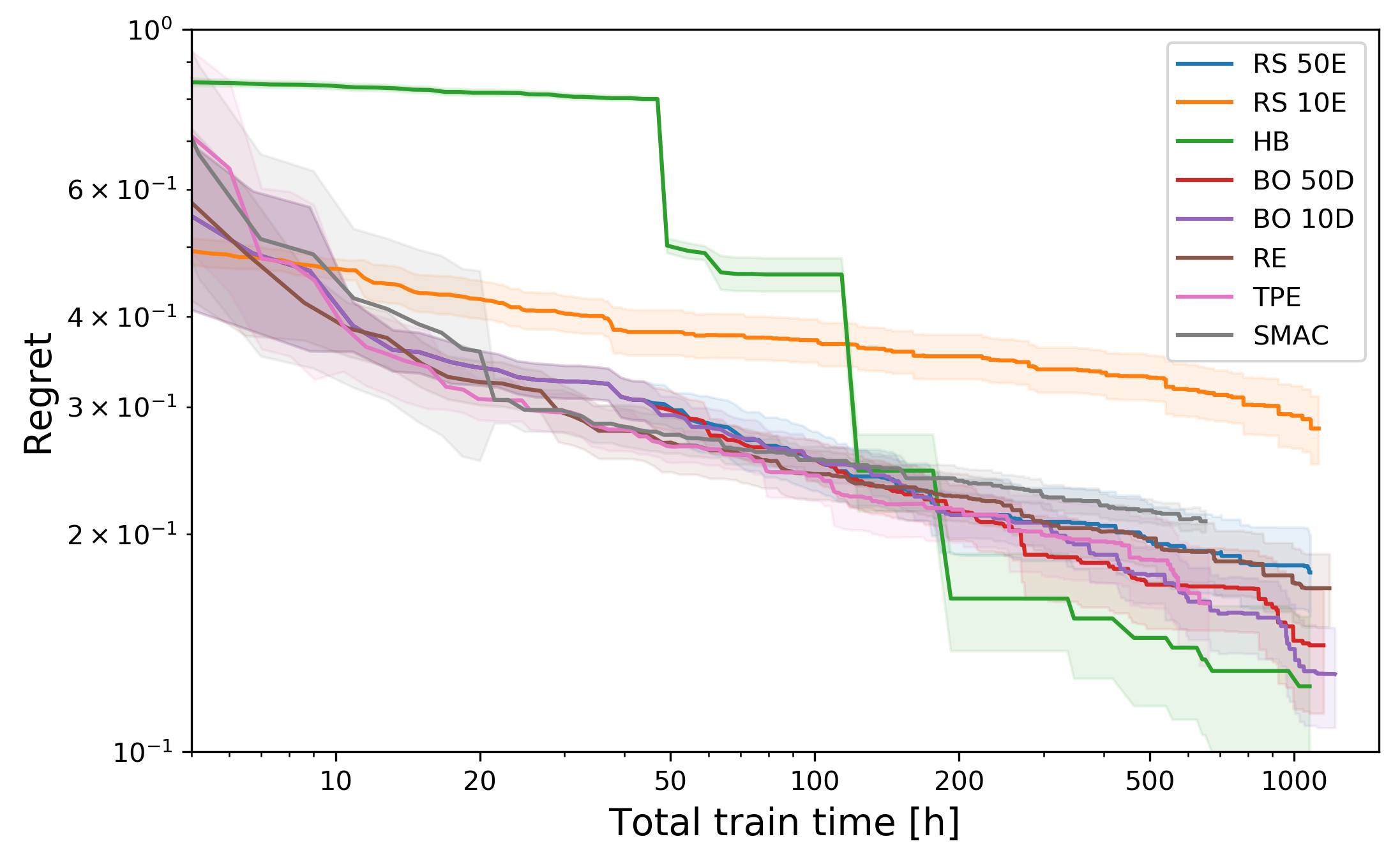}
\caption{Performance of various NAS methods. Shades of curves correspond to 95\%-confidence interval for mean values.}
\label{fig:nas_benchmark}
\vspace{-45pt}
\end{wrapfigure}
Performance of each method was measured with regret vs. total training time, where regret $r$ at the moment $t$ is $r(t) = L(t) - L^*$, $L(t)$ is the final testing log perplexity of the best architecture according to validation perplexity found so far by the moment $t$, and $L^*$ is the lowest testing log perplexity in the whole dataset (in our benchmark $L^* = 4.36$, achieved by LSTM architecture). For each method, we report the average regret over 30 trials in Figure \ref{fig:nas_benchmark}. Hyperbands achive the lowest final regret, while BO follows next.  

\vspace{-5pt}


\section{Discussion}
\label{sec:discussion}

The proposed benchmark is in a different vein than the previous ones. Firstly, it has a much more complex search space, however, at the price of being a very sparse sample from it. Secondly, as the analysis has shown, the distribution of performance metrics (perplexity) is not skewed towards the optimum as in Nas-Bench-101, moreover, hand-crafted architectures like LSTM and GRU seem to have a streamlined performance, hardly achievable by random instances from the search space. We believe these peculiarities of our benchmark will bring diversity and new challenges to the neural architecture search community. For example, larger architectures, that are more realistic, pose a challenge on feature engineering, since simple approaches like flat encoding of adjacency matrices of architectures' graphs would suffer from the curse of dimensionality. In this work, we used graph2vec approach to obtain better features in a small dimensional space, but we leave a space for further experiments with graph neural networks and other graph-encoding techniques.

\vspace{-5pt}
\section{Conclusion}
\label{sec:conclusion}

In this work, we introduced a novel benchmark for the search of recurrent neural architectures for language modeling. The complexity of recurrent cells opens new opportunities to experimenting with sophisticated feature engineering methods. With the data we generated, we have found that the performance of architectures highly correlates on different datasets. The results also extend the findings of previous works, that GRU and LSTM architectures generally have the top performance among others. While previously such conclusions were made based on analysis of local neighborhoods of those architectures, our work confirms them on a global scale; however, we have also found a few different architectures with the similar performance. We hope this benchmark will bring new insights regarding the performance of various recurrent architectures and better NAS methods.  
\section{Acknowledgements}
\label{sec:acknowledgements}

This work was done during the cooperation project with Huawei Noah's Ark Lab. 

We thank Alexander Filippov from Huawei Noah's Ark Lab for discussion of problem statements and comments about industrial applications of NAS. 

We acknowledge the usage of the Skoltech CDISE HPC cluster Zhores for obtaining the results presented in this paper.

Ekaterina Artemova was supported by the framework of the HSE University Basic Research Program funded by the Russian Academic Excellence Project ‘5-100’. Her research was carried out on HPC facilities at NRU HSE.


\bibliographystyle{abbrv}
\bibliography{bibliography}

\begin{thebibliography}{10}

\bibitem{awd_lstm_repo}
Lstm and qrnn language model toolkit.
\newblock \url{https://github.com/salesforce/awd-lstm-lm}.
\newblock Accessed: 2020-06-02.

\bibitem{abu2015exact}
Z.~Abu-Aisheh, R.~Raveaux, J.-Y. Ramel, and P.~Martineau.
\newblock An exact graph edit distance algorithm for solving pattern
  recognition problems.
\newblock In {\em 4th International Conference on Pattern Recognition
  Applications and Methods 2015}, 2015.

\bibitem{baroni2014don}
M.~Baroni, G.~Dinu, and G.~Kruszewski.
\newblock Don’t count, predict! a systematic comparison of context-counting
  vs. context-predicting semantic vectors.
\newblock In {\em Proceedings of the 52nd Annual Meeting of the Association for
  Computational Linguistics (Volume 1: Long Papers)}, pages 238--247, 2014.

\bibitem{bengio2003neural}
Y.~Bengio, R.~Ducharme, P.~Vincent, and C.~Jauvin.
\newblock A neural probabilistic language model.
\newblock {\em Journal of machine learning research}, 3(Feb):1137--1155, 2003.

\bibitem{bergstra2013making}
J.~Bergstra, D.~Yamins, and D.~D. Cox.
\newblock Making a science of model search: Hyperparameter optimization in
  hundreds of dimensions for vision architectures.
\newblock 2013.

\bibitem{breiman1996bagging}
L.~Breiman.
\newblock Bagging predictors.
\newblock {\em Machine learning}, 24(2):123--140, 1996.

\bibitem{britz2017massive}
D.~Britz, A.~Goldie, M.-T. Luong, and Q.~Le.
\newblock Massive exploration of neural machine translation architectures.
\newblock In {\em Proceedings of the 2017 Conference on Empirical Methods in
  Natural Language Processing}, pages 1442--1451, 2017.

\bibitem{chen2016xgboost}
T.~Chen and C.~Guestrin.
\newblock Xgboost: A scalable tree boosting system.
\newblock In {\em Proceedings of the 22nd acm sigkdd international conference
  on knowledge discovery and data mining}, pages 785--794, 2016.

\bibitem{cho2014properties}
K.~Cho, B.~Van~Merri{\"e}nboer, D.~Bahdanau, and Y.~Bengio.
\newblock On the properties of neural machine translation: Encoder-decoder
  approaches.
\newblock {\em arXiv preprint arXiv:1409.1259}, 2014.

\bibitem{clark2019electra}
K.~Clark, M.-T. Luong, Q.~V. Le, and C.~D. Manning.
\newblock Electra: Pre-training text encoders as discriminators rather than
  generators.
\newblock In {\em International Conference on Learning Representations}, 2019.

\bibitem{Dong2020NAS-Bench-201:}
X.~Dong and Y.~Yang.
\newblock Nas-bench-201: Extending the scope of reproducible neural
  architecture search.
\newblock In {\em International Conference on Learning Representations}, 2020.

\bibitem{finkelstein2001placing}
L.~Finkelstein, E.~Gabrilovich, Y.~Matias, E.~Rivlin, Z.~Solan, G.~Wolfman, and
  E.~Ruppin.
\newblock Placing search in context: The concept revisited.
\newblock In {\em Proceedings of the 10th international conference on World
  Wide Web}, pages 406--414, 2001.

\bibitem{frazier2018tutorial}
P.~I. Frazier.
\newblock A tutorial on bayesian optimization.
\newblock {\em arXiv preprint arXiv:1807.02811}, 2018.

\bibitem{greff2016lstm}
K.~Greff, R.~K. Srivastava, J.~Koutn{\'\i}k, B.~R. Steunebrink, and
  J.~Schmidhuber.
\newblock Lstm: A search space odyssey.
\newblock {\em IEEE transactions on neural networks and learning systems},
  28(10):2222--2232, 2016.

\bibitem{hagberg2008exploring}
A.~Hagberg, P.~Swart, and D.~S~Chult.
\newblock Exploring network structure, dynamics, and function using networkx.
\newblock Technical report, Los Alamos National Lab.(LANL), Los Alamos, NM
  (United States), 2008.

\bibitem{he2016deep}
K.~He, X.~Zhang, S.~Ren, and J.~Sun.
\newblock Deep residual learning for image recognition.
\newblock In {\em Proceedings of the IEEE conference on computer vision and
  pattern recognition}, pages 770--778, 2016.

\bibitem{hill2015simlex}
F.~Hill, R.~Reichart, and A.~Korhonen.
\newblock Simlex-999: Evaluating semantic models with (genuine) similarity
  estimation.
\newblock {\em Computational Linguistics}, 41(4):665--695, 2015.

\bibitem{hochreiter1997long}
S.~Hochreiter and J.~Schmidhuber.
\newblock Long short-term memory.
\newblock {\em Neural computation}, 9(8):1735--1780, 1997.

\bibitem{huang2017densely}
G.~Huang, Z.~Liu, L.~Van Der~Maaten, and K.~Q. Weinberger.
\newblock Densely connected convolutional networks.
\newblock In {\em Proceedings of the IEEE conference on computer vision and
  pattern recognition}, pages 4700--4708, 2017.

\bibitem{hutter2011sequential}
F.~Hutter, H.~H. Hoos, and K.~Leyton-Brown.
\newblock Sequential model-based optimization for general algorithm
  configuration.
\newblock In {\em International conference on learning and intelligent
  optimization}, pages 507--523. Springer, 2011.

\bibitem{jozefowicz2016exploring}
R.~Jozefowicz, O.~Vinyals, M.~Schuster, N.~Shazeer, and Y.~Wu.
\newblock Exploring the limits of language modeling.
\newblock {\em arXiv preprint arXiv:1602.02410}, 2016.

\bibitem{jozefowicz2015empirical}
R.~Jozefowicz, W.~Zaremba, and I.~Sutskever.
\newblock An empirical exploration of recurrent network architectures.
\newblock In {\em International conference on machine learning}, pages
  2342--2350, 2015.

\bibitem{kenton2019bert}
J.~D. M.-W.~C. Kenton and L.~K. Toutanova.
\newblock Bert: Pre-training of deep bidirectional transformers for language
  understanding.
\newblock In {\em Proceedings of NAACL-HLT}, pages 4171--4186, 2019.

\bibitem{klein2019tabular}
A.~Klein and F.~Hutter.
\newblock Tabular benchmarks for joint architecture and hyperparameter
  optimization.
\newblock {\em arXiv preprint arXiv:1905.04970}, 2019.

\bibitem{li2017hyperband}
L.~Li, K.~Jamieson, G.~DeSalvo, A.~Rostamizadeh, and A.~Talwalkar.
\newblock Hyperband: A novel bandit-based approach to hyperparameter
  optimization.
\newblock {\em The Journal of Machine Learning Research}, 18(1):6765--6816,
  2017.

\bibitem{li2019random}
L.~Li and A.~Talwalkar.
\newblock Random search and reproducibility for neural architecture search.
\newblock {\em arXiv preprint arXiv:1902.07638}, 2019.

\bibitem{lindauer2019best}
M.~Lindauer and F.~Hutter.
\newblock Best practices for scientific research on neural architecture search.
\newblock {\em arXiv preprint arXiv:1909.02453}, 2019.

\bibitem{liu2018darts}
H.~Liu, K.~Simonyan, and Y.~Yang.
\newblock Darts: Differentiable architecture search.
\newblock {\em arXiv preprint arXiv:1806.09055}, 2018.

\bibitem{merity2017regularizing}
S.~Merity, N.~S. Keskar, and R.~Socher.
\newblock Regularizing and optimizing lstm language models.
\newblock {\em arXiv preprint arXiv:1708.02182}, 2017.

\bibitem{merity2016pointer}
S.~Merity, C.~Xiong, J.~Bradbury, and R.~Socher.
\newblock Pointer sentinel mixture models.
\newblock {\em arXiv preprint arXiv:1609.07843}, 2016.

\bibitem{mikolov2010recurrent}
T.~Mikolov, M.~Karafi{\'a}t, L.~Burget, J.~{\v{C}}ernock{\`y}, and
  S.~Khudanpur.
\newblock Recurrent neural network based language model.
\newblock In {\em Eleventh annual conference of the international speech
  communication association}, 2010.

\bibitem{mikolov2013distributed}
T.~Mikolov, I.~Sutskever, K.~Chen, G.~S. Corrado, and J.~Dean.
\newblock Distributed representations of words and phrases and their
  compositionality.
\newblock In {\em Advances in neural information processing systems}, pages
  3111--3119, 2013.

\bibitem{narayanan2017graph2vec}
A.~Narayanan, M.~Chandramohan, R.~Venkatesan, L.~Chen, Y.~Liu, and S.~Jaiswal.
\newblock graph2vec: Learning distributed representations of graphs.
\newblock {\em arXiv preprint arXiv:1707.05005}, 2017.

\bibitem{pham2018efficient}
H.~Pham, M.~Y. Guan, B.~Zoph, Q.~V. Le, and J.~Dean.
\newblock Efficient neural architecture search via parameter sharing.
\newblock {\em arXiv preprint arXiv:1802.03268}, 2018.

\bibitem{ponte1998language}
J.~M. Ponte and W.~B. Croft.
\newblock A language modeling approach to information retrieval.
\newblock In {\em Proceedings of the 21st annual international ACM SIGIR
  conference on Research and development in information retrieval}, pages
  275--281, 1998.

\bibitem{press2017using}
O.~Press and L.~Wolf.
\newblock Using the output embedding to improve language models.
\newblock In {\em Proceedings of the 15th Conference of the European Chapter of
  the Association for Computational Linguistics: Volume 2, Short Papers}, pages
  157--163, 2017.

\bibitem{raffel2019exploring}
C.~Raffel, N.~Shazeer, A.~Roberts, K.~Lee, S.~Narang, M.~Matena, Y.~Zhou,
  W.~Li, and P.~J. Liu.
\newblock Exploring the limits of transfer learning with a unified text-to-text
  transformer.
\newblock {\em arXiv preprint arXiv:1910.10683}, 2019.

\bibitem{real2019regularized}
E.~Real, A.~Aggarwal, Y.~Huang, and Q.~V. Le.
\newblock Regularized evolution for image classifier architecture search.
\newblock In {\em Proceedings of the aaai conference on artificial
  intelligence}, volume~33, pages 4780--4789, 2019.

\bibitem{schnabel2015evaluation}
T.~Schnabel, I.~Labutov, D.~Mimno, and T.~Joachims.
\newblock Evaluation methods for unsupervised word embeddings.
\newblock In {\em Proceedings of the 2015 conference on empirical methods in
  natural language processing}, pages 298--307, 2015.

\bibitem{tay2018densely}
Y.~Tay, A.~T. Luu, S.~C. Hui, and J.~Su.
\newblock Densely connected attention propagation for reading comprehension.
\newblock In {\em Advances in Neural Information Processing Systems}, pages
  4906--4917, 2018.

\bibitem{wang2018glue}
A.~Wang, A.~Singh, J.~Michael, F.~Hill, O.~Levy, and S.~Bowman.
\newblock Glue: A multi-task benchmark and analysis platform for natural
  language understanding.
\newblock In {\em Proceedings of the 2018 EMNLP Workshop BlackboxNLP: Analyzing
  and Interpreting Neural Networks for NLP}, pages 353--355, 2018.

\bibitem{wang2019jiant}
A.~Wang, I.~F. Tenney, Y.~Pruksachatkun, P.~Yeres, J.~Phang, H.~Liu, P.~M.
  Htut, , K.~Yu, J.~Hula, P.~Xia, R.~Pappagari, S.~Jin, R.~T. McCoy, R.~Patel,
  Y.~Huang, E.~Grave, N.~Kim, T.~F\'evry, B.~Chen, N.~Nangia, A.~Mohananey,
  K.~Kann, S.~Bordia, N.~Patry, D.~Benton, E.~Pavlick, and S.~R. Bowman.
\newblock \texttt{jiant} 1.3: A software toolkit for research on
  general-purpose text understanding models.
\newblock \url{http://jiant.info/}, 2019.

\bibitem{wang2018densely}
S.~Wang, M.~Huang, and Z.~Deng.
\newblock Densely connected cnn with multi-scale feature attention for text
  classification.
\newblock In {\em IJCAI}, pages 4468--4474, 2018.

\bibitem{yang2019evaluation}
A.~Yang, P.~M. Esperan{\c{c}}a, and F.~M. Carlucci.
\newblock Nas evaluation is frustratingly hard.
\newblock {\em arXiv preprint arXiv:1912.12522}, 2019.

\bibitem{ying2019bench}
C.~Ying, A.~Klein, E.~Christiansen, E.~Real, K.~Murphy, and F.~Hutter.
\newblock Nas-bench-101: Towards reproducible neural architecture search.
\newblock In {\em International Conference on Machine Learning}, pages
  7105--7114, 2019.

\bibitem{zacharov2019zhores}
I.~Zacharov, R.~Arslanov, M.~Gunin, D.~Stefonishin, A.~Bykov, S.~Pavlov,
  O.~Panarin, A.~Maliutin, S.~Rykovanov, and M.~Fedorov.
\newblock “zhores”—petaflops supercomputer for data-driven modeling,
  machine learning and artificial intelligence installed in skolkovo institute
  of science and technology.
\newblock {\em Open Engineering}, 9(1):512--520.

\bibitem{zela2020bench}
A.~Zela, J.~Siems, and F.~Hutter.
\newblock Nas-bench-1shot1: Benchmarking and dissecting one-shot neural
  architecture search.
\newblock {\em arXiv preprint arXiv:2001.10422}, 2020.

\bibitem{zilly2017recurrent}
J.~G. Zilly, R.~K. Srivastava, J.~Koutn{\'\i}k, and J.~Schmidhuber.
\newblock Recurrent highway networks.
\newblock In {\em Proceedings of the 34th International Conference on Machine
  Learning-Volume 70}, pages 4189--4198. JMLR. org, 2017.

\bibitem{zoph2016neural}
B.~Zoph and Q.~V. Le.
\newblock Neural architecture search with reinforcement learning.
\newblock {\em arXiv preprint arXiv:1611.01578}, 2016.

\end{thebibliography}

\clearpage
\appendix
\section{The top architectures}
\label{sec:appendix_top_archs}

\begin{figure}[!ht]
\centering
\begin{subfigure}[b]{.46\linewidth} 
\includegraphics[width=0.80\textwidth]{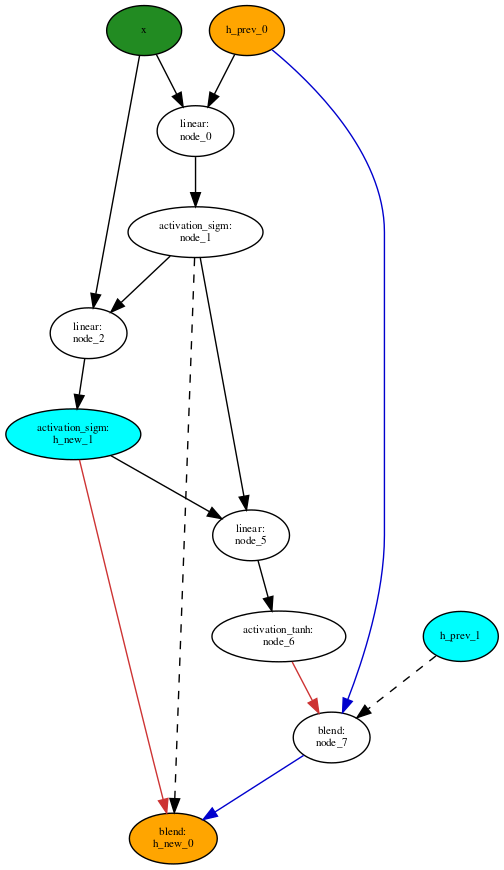}
\caption{Top-2}
\end{subfigure}
\begin{subfigure}[b]{.46\linewidth} 
\includegraphics[width=0.99\textwidth]{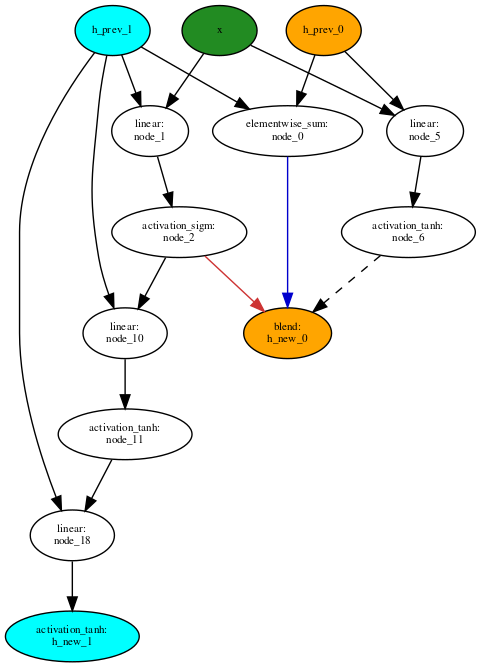}
\caption{Top-4}
\end{subfigure}
\caption{The top non-conventional architectures according to test perplexity.}
\label{fig:best_archs}
\end{figure}

Figure \ref{fig:rnn_cells} already contains two top architectures - LSTM and GRU. Two more randomly generated architectures with competitive performance (test perplexity) are provided in Figure \ref{fig:best_archs}; they are substantially different from LSTM and GRU.

\section{Additional analysis of NAS methods}
\label{sec:appendix_additional_analysis}
\begin{figure}[ht]
\centering
\includegraphics[width=0.6\textwidth]{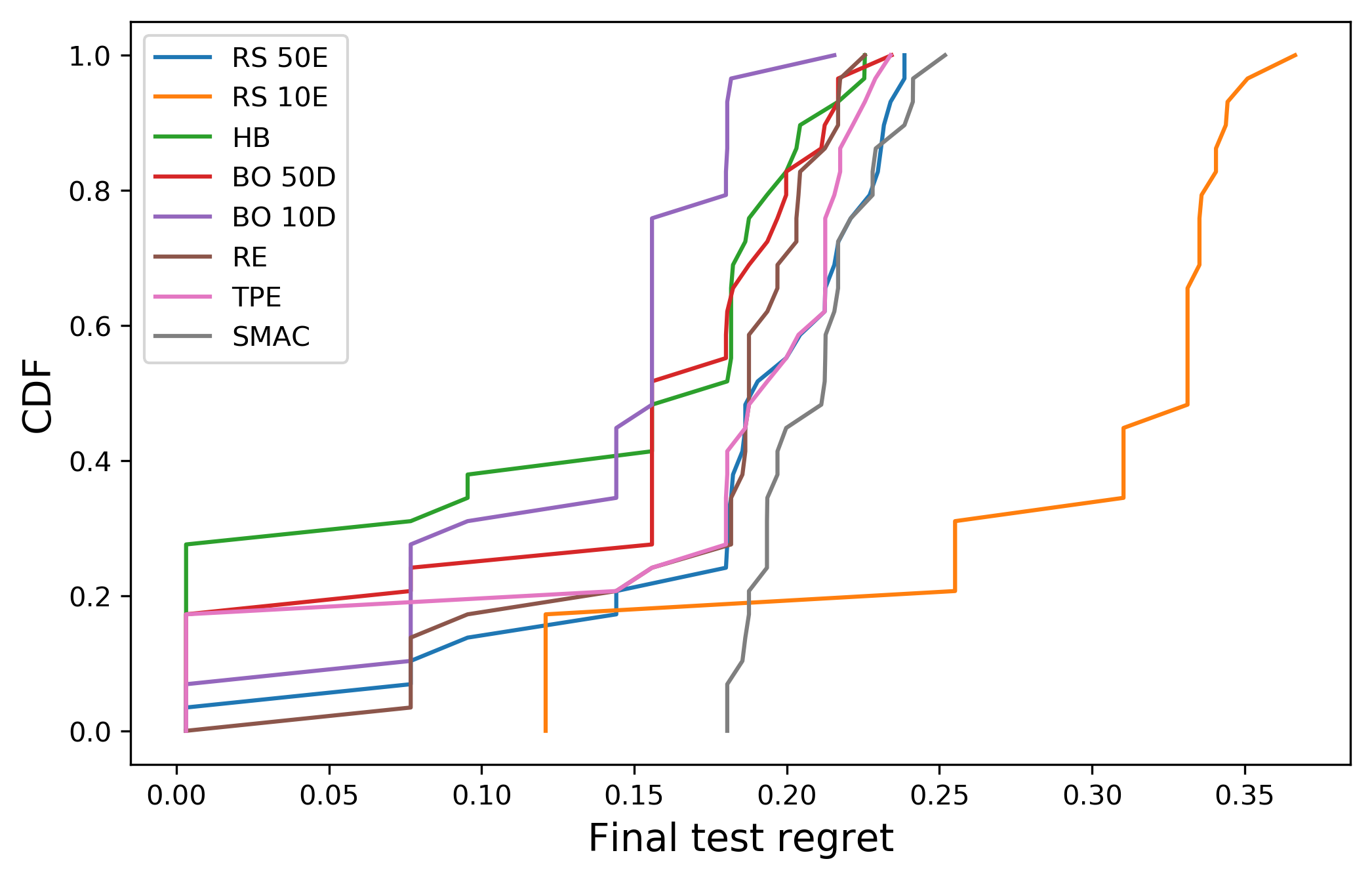}
\caption{Distribution of the final testing regrets w.r.t. various seeds.}
\label{fig:cdf-regret}
\end{figure}

Figure \ref{fig:cdf-regret} is complementary to Figure \ref{fig:nas_benchmark}, it shows the cumulative distribution of the final testing regrets among multiple random seeds. BO 50D, TPE and HB manage to find the best architecture (regret = 0) within 1000 hours in approximately 20\% runs. 

\section{Extended intrinsic evaluation}
\begin{figure}[ht]
\centering
\begin{subfigure}{.46\textwidth} 
\includegraphics[width=\textwidth]{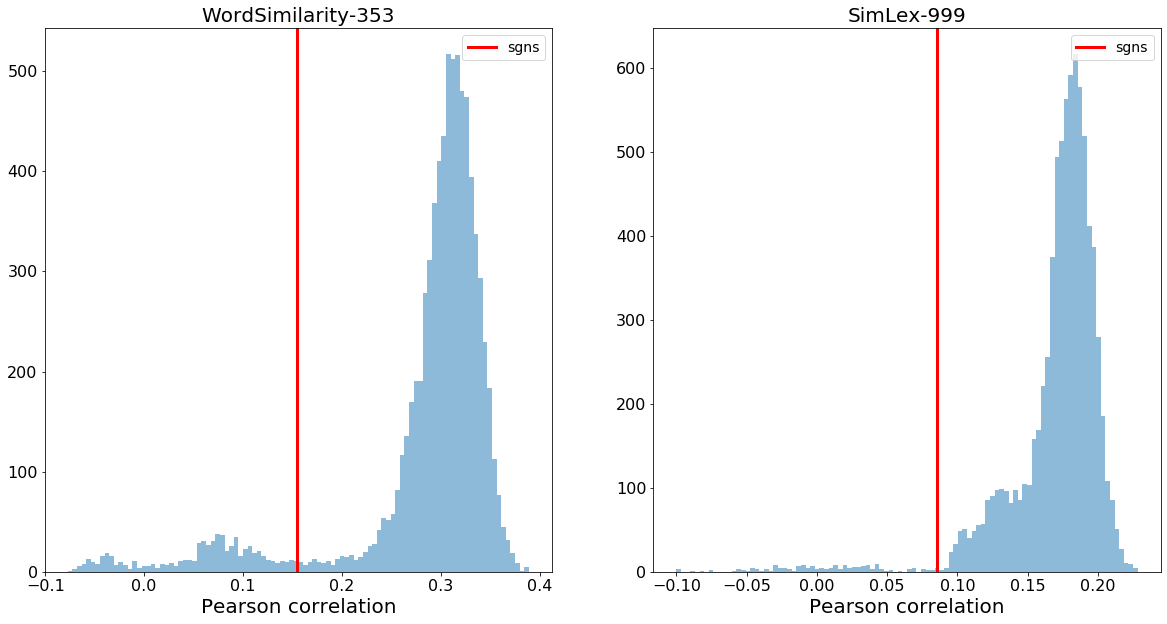}
\caption{}
\end{subfigure}
\begin{subfigure}{.46\textwidth} 
\includegraphics[width=\textwidth]{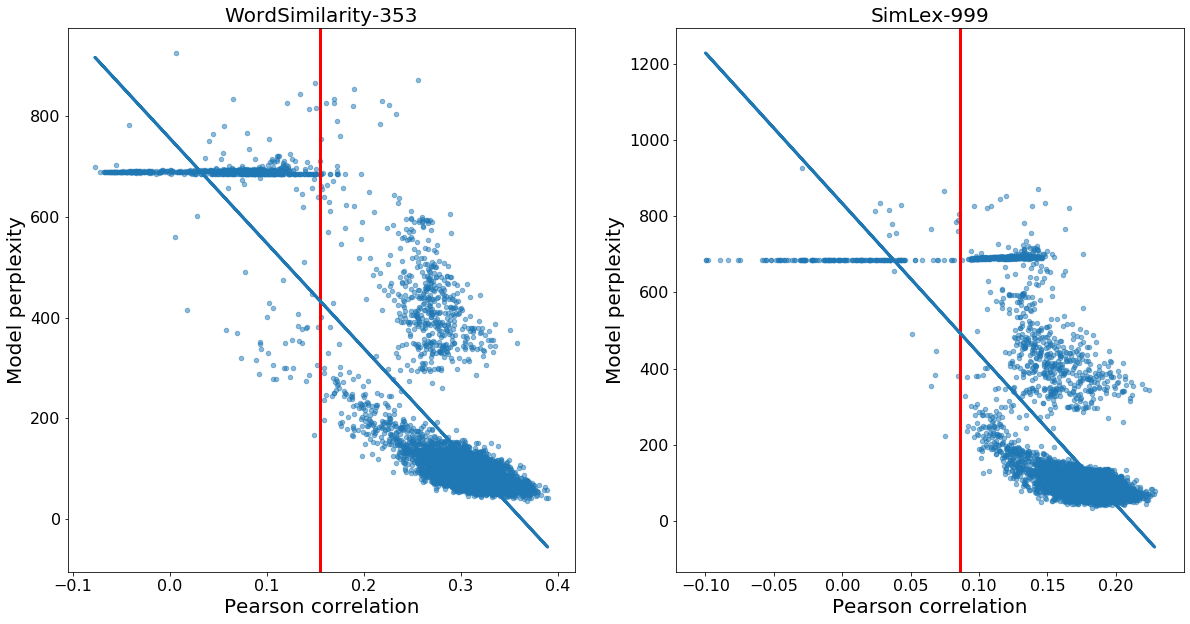}
\caption{}
\label{fig:static_we_eval_ext_scatter}
\end{subfigure}
\setlength{\belowcaptionskip}{-10pt}
\caption{Left: Histograms of WordSimilarity-353 and SimLex-999. Right: Scatter-plots of WordSimilarity-353 and SimLex-999 vs. perplexity. WordSimilarity-353 and SimLex-999 are measured by Pearson correlation coefficient}
\label{fig:static_we_eval_ext}
\end{figure}

Figure \ref{fig:static_we_eval_ext} shows metrics of the intrinsic evaluation of all the 14k architectures (trained on PTB) from the search space. Figure \ref{fig:static_we_eval_ext_scatter} is an extension of the Figure \ref{fig:static_we_eval}, the latter one was calculated based on a stratified sample of the architectures. The red baseline indicates SGNS performance. Most of the architectures overcome this baseline. There are interesting clusters of architectures on Figure \ref{fig:static_we_eval_ext_scatter} that have high perplexity (around 400), yet almost the same correlation scores as architectures with low perplexity (0.15 on average). Such cases can take place due to degenerate cell structures, that do not allow networks to learn predictions well.

\section{Detailed extrinsic evaluation}
Table \ref{tbl:det_extr_eval} shows detailed results of testing 150 random architectures, trained on Wikitext-2, on GLUE benchmark. The results are lower than the baseline, adopted from \cite{wang2018glue}.  The possible explanation for this is due to the vocabulary size. The baseline, which has been trained on the GLUE datasets from scratch, thus possess all GLUE vocabulary. The architectures are pretrained on WikiText-2, hence their vocabulary omits part of the GLUE vocabulary. For some simpler paraphrase tasks, such as QQP and MRPC, the architectures get close to baselines. This evidences the architectures are capable for solving tasks, that require understanding of semantics, but fail to capture more complex phenomena, such as natural language inference (NLI and RTE tasks) and language acceptability (CoLA).

\begin{table}[ht]
\centering
\caption{GLUE scores on separate tasks. Score shows the average metric on all tasks multiplied by 100.}
\label{tbl:det_extr_eval}
\begin{tabular}{ccccccc}
\toprule
 & mean  & std  & min  & median  & max & baseline \cite{wang2018glue} \\
\midrule
cola\_mcc  & 0.000 & 0.005 & -0.046 & 0.000 & 0.025 & 0.35 \\
sst\_accuracy  & 0.524 & 0.039 & 0.471 & 0.509 & 0.638 & 0.9 \\
mrpc\_f1  & 0.805 & 0.036 & 0.501 & 0.812 & 0.817 & 0.84 \\
mrpc\_accuracy  & 0.680 & 0.024 & 0.478 & 0.684 & 0.711 & 0.78 \\
sts-b\_pearsonr  & 0.001 & 0.076 & -0.166 & -0.004 & 0.309 & 0.79 \\
sts-b\_spearmanr  & -0.001 & 0.076 & -0.164 & -0.010 & 0.312 & 0.79 \\
qqp\_f1  & 0.532 & 0.086 & 0.000 & 0.542 & 0.613 & 0.66\\
qqp\_accuracy  & 0.488 & 0.117 & 0.368 & 0.437 & 0.668 & 0.865 \\
mnli\_accuracy  & 0.363 & 0.025 & 0.328 & 0.360 & 0.438 & 0.769 \\
qnli\_accuracy  & 0.503 & 0.012 & 0.494 & 0.496 & 0.545 & 0.798 \\
rte\_accuracy  & 0.527 & 0.000 & 0.527 & 0.527 & 0.527 & 0.592 \\
wnli\_accuracy  & 0.527 & 0.058 & 0.338 & 0.563 & 0.634 & 0.651 \\
glue-diagnostic\_all\_mcc  & 0.003 & 0.030 & -0.057 & 0.000 & 0.094 & 0.28 \\
score & 38.1 & 1.8 & 34.7 & 37.9 & 43.6 & 69.1 \\

\bottomrule
\end{tabular}
\end{table}



\end{document}